\documentclass{article}

\usepackage{longtable}
\usepackage{array}
\usepackage{booktabs}

\usepackage{PRIMEarxiv}
\usepackage{longtable}
\usepackage{amsmath}
\usepackage{svg}
\usepackage[utf8]{inputenc} % allow utf-8 input
\usepackage[T1]{fontenc}    % use 8-bit T1 fonts
\usepackage{hyperref}       % hyperlinks
\usepackage{url}            % simple URL typesetting
\usepackage{booktabs}       % professional-quality tables
\usepackage{amsfonts}       % blackboard math symbols
\usepackage{nicefrac}       % compact symbols for 1/2, etc.
\usepackage{microtype}      % microtypography
\usepackage{lipsum}
\usepackage{fancyhdr}       % header
\usepackage{graphicx}       % graphics
\graphicspath{{media/}}     % organize your images and other figures under media/ folder
\usepackage{array,multirow}
\usepackage{float}
\usepackage{placeins}

\newcommand{\beginsupplement}{%
        \setcounter{table}{0}
        \renewcommand{\thetable}{S\arabic{table}}%
        \renewcommand{\theHtable}{S\arabic{table}}%
        \setcounter{figure}{0}
        \renewcommand{\thefigure}{S\arabic{figure}}%
        \renewcommand{\theHfigure}{S\arabic{figure}}%
     }

\title{LLM-Augmented Therapy Normalization and Aspect-Based Sentiment Analysis for Treatment-Resistant Depression on Reddit}
\author{
  Yuxin Zhu\\
  Department of Biomedical Informatics \\
  Emory University \\
  Atlanta, GA\\
  \texttt{yuxin.zhu@emory.edu} \\
   \And
  Sahithi Lakamana\\
  Department of Biomedical Informatics \\
  Emory University \\
  Atlanta\\
  \texttt{sahithi.krishnaveni.lakamana@emory.edu} \\
  \And
  Masoud Rouhizadeh\\
  Department of Pharmaceutical Outcomes and Policy \\
  University of Florida \\
  Gainesville, FL\\
  \texttt{mrouhizadeh@ufl.edu} \\
   \And
   Selen Bozkurt\\
  Department of Biomedical Informatics \\
  Emory University \\
  Atlanta, GA \\
  \texttt{selen.bozkurt@emory.edu} \\
  \And
  Rachel Hershenberg\\
  Department of Psychiatry and Behavioral Sciences \\
  Emory University \\
  Atlanta, GA \\
  \texttt{rachel.hershenberg@emory.edu} \\
   \And
  Abeed Sarker\\
  Department of Biomedical Informatics \\
  Emory University \\
  Atlanta, GA\\
  \texttt{abeed@dbmi.emory.edu} \\
}

\begin{document}
\maketitle

\begin{abstract}
Treatment-resistant depression (TRD) is a severe form of major depressive disorder in which patients do not achieve remission despite multiple adequate treatment trials. Evidence across pharmacologic options for TRD remains limited, and trials often incompletely reflect patient-reported tolerability. Large-scale online peer-support narratives therefore offer a complementary lens on how patients describe and evaluate medications in real-world use. In this study, we curated a corpus of 5,059 Reddit posts explicitly referencing TRD from 3,480 subscribers across 28 mental health–related subreddits (2010–2025). From these, 3,839 posts mentioning at least one medication were identified, yielding 23,399 mentions of 81 generic-name medications after lexicon-based normalization of brand names, misspellings, and colloquialisms. We developed an aspect-based sentiment classifier by fine-tuning a DeBERTa-v3 model on the SMM4H 2023 therapy-sentiment Twitter corpus with large language model–based data augmentation, achieving a micro-$F_1$ of 0.800 on the shared task test set. Applying this classifier to Reddit, we quantified sentiment toward individual medications across three dimensions (positive, neutral, negative) and tracked patterns by drug, subscriber, subreddit, and year. Overall, 72.1\% of medication mentions were neutral, 14.8\% negative, and 13.1\% positive. Conventional antidepressants (SSRIs and SNRIs) showed consistently higher negative than positive proportions, whereas ketamine and esketamine exhibited comparatively more favorable sentiment profiles, with positive mentions approximating or slightly exceeding negative mentions. These findings demonstrate that normalized medication extraction combined with aspect-based sentiment analysis can help characterize patient-perceived treatment experiences in unstructured TRD-related Reddit discourse, complementing clinical evidence with large-scale, patient-generated perspectives.
\end{abstract}

% keywords can be removed

\section{Introduction}

Treatment-resistant depression (TRD) is a severe subtype of major depressive disorder (MDD) characterized by a lack of response to multiple standard antidepressant treatment regimens \cite{mcintyre2023treatment,doi:10.1176/ajp.2006.163.11.1905,otte2016major}. While MDD can often be managed with therapy and medication, approximately 30\% of patientes will be considered treatment resistant, often defined as a failure to achieve remission, despite at least two adequately dosed  antidepressant trials \cite{cipriani2018comparative,https://doi.org/10.1002/prp2.472,parikh2004current,ruhe2012staging,mcintyre2014treatment}. TRD patients carry a disproportionately high burden of illness, relative to their treatment responsive counterparts, including greater functional impairment \cite{vancappel2021cognitive,gregory2020predictors,judd2000psychosocial,thase1994refractory}, lower quality of life \cite{lex2019quality,rathod2022health}, increased risk of comorbidities \cite{adekkanattu2023comorbidity} and suicidality \cite{souery2007clinical}. The burden of TRD extends beyond individual suffering to substantial societal and healthcare costs; in the United States, TRD has been estimated to affect millions of adults and to account for a large share of depression-related economic burden \cite{zhdanava2021prevalence}. These realities have driven expansion of interventional psychiatry beyond conventional antidepressants, encompassing neuromodulation (e.g., ECT, TMS, VNS) \cite{cusin2012somatic} and newer pharmacotherapies such as ketamine, esketamine, alongside emerging investigational agents including psychedelics \cite{d2025psychedelics,ng2021efficacy}. Although controlled trials and meta-analyses increasingly quantify overall response and remission rates for these interventions \cite{berlim2014response,nikolin2023ketamine,george2010daily}, the evidence base remains comparatively nascent with respect to moderators of response and patient-centered outcomes \cite{price2022international}. However, clinical evidence is limited and does not fully capture how patients experience these therapies in everyday life, including perceived benefit, tolerability, and practical barriers such as access and adherence. This type of highly generalizable, patient-centered data can complement more tightly controlled clinical efficacy data.

Indeed, patient-generated health data, including spontaneously produced narratives on social media, are a source of real-world insight that can complement conventional clinical evidence \cite{cimiano2024patient}. Individuals often describe medication use (both medical and nonmedical), side effects, discontinuation, and switching decisions in their own words on social media \cite{wicks2010sharing,golder2023reasons,golder2022patient}. These narratives can reveal concerns that are difficult to measure consistently in clinical settings, and at a scale that is challenging to obtain through traditional surveys or cohorts \cite{bunting2021socially,saha2021understanding,frost2011patient,wicks2011accelerated}. Reddit is particularly well-suited for studying stigmatized and sensitive health discussions because it is organized into topic-focused communities (subreddits) and supports pseudonymous participation. Reddit's pseudonymous design can reduce barriers to disclosure and facilitate candid discussion of sensitive health topics, such as conditions where sharing experiences may carry social, reputational, or psychological risk, including but not limited to depression and suicidality, eating disorders, substance use, and sexual health concerns \cite{de2014mental,pavalanathan2015identity,ammari2019self, sowles2018content,kepner2022types,chi2023investigating,nobles2018std}. In mental health communities specifically, De Choudhury and Kiciman \cite{de2017language} reported that the language of social support exchanged in Reddit forums is associated with subsequent suicidal ideation risk, demonstrating both the candor and clinical relevance of these peer-support settings. Consistent with this broader view, prior work has characterized discussion themes and dynamics across large Reddit mental health communities \cite{park2018examining,low2020natural,grub2025reddit}. A growing body of work further demonstrates that natural language processing (NLP) can model mental health conditions and risk signals from social media language using large, public datasets \cite{thorstad2019predicting,coppersmith2018natural,cohan2018smhd,jiang2020detection}. Social media has also been studied as a resource for medication-related surveillance, including pharmacovigilance, since subscribers often report side effects, discontinuation reasons, and experiential trade-offs in their own language \cite{magge2021deepademiner,golder2024value}. In the context of TRD, in which patients typically explore multiple therapies, the content of online discussions is not only about symptoms but also broader treatment evaluation: wperceived efficacy and treatment failure. Compared with the general depression literature, TRD-specific analyses that compare patient sentiments across a wide range of therapies remain limited, despite the clinical importance of understanding how patients experience an evolving treatment landscape. Systematically quantifying TRD patient sentiments from social media chatter remains an unaddressed research gap.

Systematically addressing this gap requires converting unstructured patient narratives into structured, medication-level evaluations, which introduces several NLP challenges.
From an NLP standpoint, converting TRD-related discussions into therapy-level evaluations requires two primary steps (i) robust medication mention detection and normalization, and (ii) medication-specific sentiment inference. Medication names on social media exhibit high lexical variability due to brand/generic variations, abbreviations, misspellings, and community-specific variants (e.g., street names), which can cause undercounting if extraction relies on exact matching \cite{lavertu2019redmed,SARKER201898,dirkson2019data}. Following accurate medication mention detection, sentiment must be inferred with respect to a specific medication mention based on local context rather than assigning a single polarity to an entire post, aligning naturally with target-dependent/aspect-level sentiment modeling \cite{GUO2023109618,ahmed2023breaking}. Finally, there is no publicly available TRD-specific corpus for sentiment, and existing TRD social media studies are qualitative or descriptive, so development of a sentiment analysis approach needs to rely on related annotated therapy-sentiment data and innovative data augmentation strategies involving synthetic data \cite{talbot2023treatmentresistantdepression,aragon2025adapting,cegin2025llms}.

In this work, we address this gap by performing aspect-based sentiment analysis of subscriber-reported evaluations of TRD therapies on Reddit. Our specific contributions are:
\begin{itemize}
\item We curate a cohort of Reddit subscribers who explicitly reference TRD and build a normalized medication representation that consolidates brand names, misspellings, and colloquialisms into generic names, enabling corpus-level characterization of medication discussions at scale.
\item We perform aspect-based sentiment classification for individual medication mentions using a transformer-based model (DeBERTa-v3) adapted from a therapy-sentiment shared-task corpus \cite{GUO2023109618,10.1093/jamia/ocae010}.

\item We employ a large language model-based data augmentation strategy to improve sentiment classification performance with synthetic data, and to address domain shift.
\item We quantify how report sentiment toward medications varies by drug, medication class, subreddit, subscriber, and year, providing a structured view of patient-reported treatment experiences that complements clinical evidence with large-scale, patient-generated perspectives.
\end{itemize}

\section{Methods}

\subsection{Data Sources and Extraction}
\begin{figure}[h!]
    \centering
    \includegraphics[width=\columnwidth]{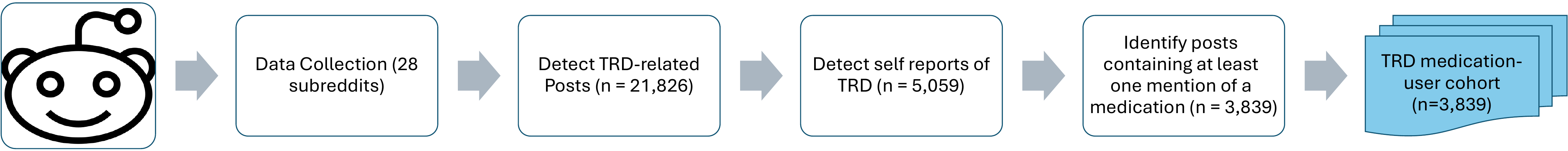}
    \caption{Overview of the data processing pipeline from raw Reddit data to the final analysis sample. Posts were first collected from 28 relevant subreddits (21,826 posts in total). TRD-related post detection yielded 5,059 posts. Filtering by medication mentions resulted in 3,839 posts. These 3,839 posts form the TRD medication-subscriber cohort analyzed in this study. Each number ($n$) represents the count of posts passing that stage.}
    \label{fig:flow}
\end{figure}

We collected posts from 28 mental health-related subreddits between 2010 and July 2025, spanning general depression communities, treatment- and medication-focused forums, and professional/advice subreddits (see Table \ref{tab:subreddits}). In total, 21{,}826 posts authored by 17{,}407 distinct subscribers were retrieved. We then applied a two-stage filtering pipeline (Figure~\ref{fig:flow}). First, TRD-relevant posts were identified using keywords (listed in Table \ref{tab:keywords}). Because the construct is variably labeled in both clinical and lay discourse, with TRD increasingly framed as difficult-to-treat depression (DTD) or treatment-refractory depression, we broadened the TRD keyword lexicon to include these alternative terms and abbreviations in addition to "TRD" and "treatment-resistant depression". These keywords captured explicit TRD mentions and descriptions of multiple treatment failures, yielding 5{,}059 posts. Second, we retained posts containing at least one medication mention from our lexicon (described below), resulting in 3{,}839 posts (17.6\% of all collected posts) contributed by 2{,}700 unique subscribers. Timestamps were recorded for temporal analyses, and subreddit names were used to group posts by community. Subscriber identifiers were used only for aggregated statistics and are anonymized in reporting. Cohort summary statistics are reported in Table~\ref{tab:trd_cohort}.

\begin{table}[ht]
\centering
\caption{Characteristics of the Reddit TRD cohort and medication annotations.}
\begin{tabular}{ll}
\hline
Characteristic & Value \\
\hline
Number of TRD posts & 5{,}059 \\
Number of unique subscribers & 3{,}480 \\
Number of subreddits & 28 \\
Time span & Feb 7, 2010 -- Jul 12, 2025 \\
Posts with at least one medication mention & 3{,}839 (75.9\%) \\
Posts with no medication mention & 1{,}220 (24.1\%) \\
Total medication mentions & 23{,}399 \\
Number of generic-name medications observed & 81 \\
Number of therapeutic classes & 16 \\
Mean medication mentions per post & 4.6 (median 3; range 0--142) \\
Mean distinct medications per post & 2.4 (median 1; range 0--40) \\
Subscribers with exactly one TRD post & 2{,}837 (81.5\%) \\
Subscribers with $\geq$ 5 TRD posts & 69 (2.0\%) \\
Subscribers with $\geq$ 10 TRD posts & 16 (0.5\%) \\
\hline
\end{tabular}
\label{tab:trd_cohort}
\end{table}

\subsection{Therapy Scope, Lexicon Construction, and Normalization}

\subsubsection{Therapy scope and inclusion criteria.}
We operationalized \emph{therapies} and refined the inclusion set iteratively by (i) grounding candidate inclusions in established TRD treatment overviews \cite{mcintyre2014treatment,murrough2025therapies} and by (ii) observation of Reddit TRD discussions to capture therapies salient to patient-reported experiences, including commonly discussed off-label agents (e.g., gabapentin, pregabalin). Consistent with TRD treatment pathways, the lexicon spans first-line antidepressants and common augmentation strategies (e.g., SSRIs, SNRIs, atypical antidepressants, antipsychotics and mood stabilizers), rapid-acting pharmacotherapies (e.g., ketamine/esketamine), and prominently discussed investigational pharmacotherapies (e.g., psilocybin). Neuromodulation was included as a distinct therapy class including ECT and TMS-family interventions (rTMS, iTBS, deep TMS). Implantable neurostimulation approaches, VNS and DBS, were excluded from the lexicon because they constitute invasive functional-neurosurgery interventions whose real-world narratives center on surgical implantation and are not well suited to a mention-level sentiment framework operating on short text windows \cite{mcintyre2023treatment,murrough2025therapies}.

\subsubsection{Lexicon construction and normalization}
To identify therapy mentions reliably, we constructed a TRD-focused lexicon beginning with a reference list of 83 treatment entities represented by generic (nonproprietary) names for medications (active ingredients; aligned where possible with International Nonproprietary Names and/or RxNorm normalized drug concepts) \cite{nelson2011normalized} and canonical names for neuromodulation modalities. For each entity, we compiled surface forms including brand names, abbreviations, common misspellings, and colloquial variants, and mapped each surface form back to its generic-name entity. The complete lexicon, comprising all observed name variants and their corresponding canonical mappings, is provided in Table \ref{tab:med}.

\begin{figure*}[ht]
\includegraphics[width=\linewidth]{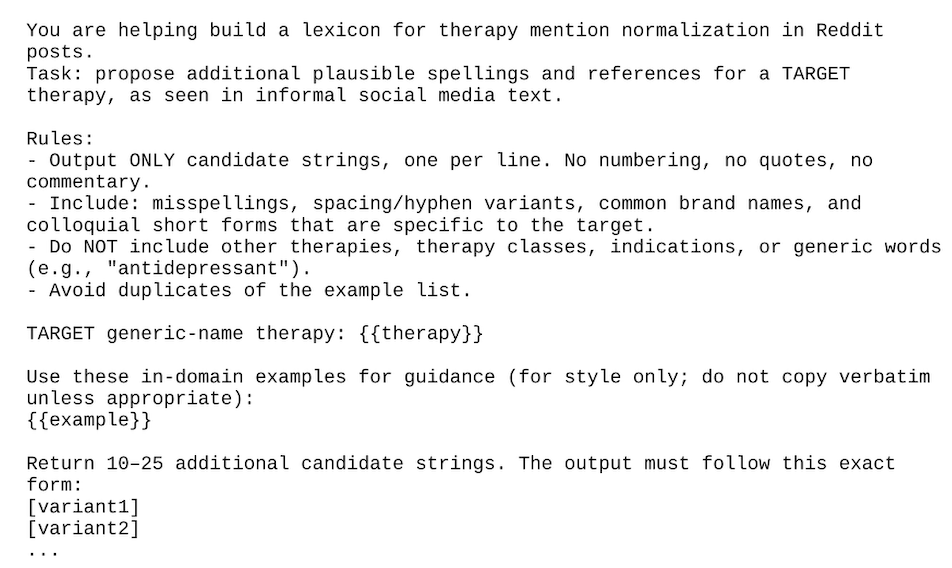}
\caption{Text of prompt used for LLM-based lexicon augmentation. \{\{therapy\}\} is replaced with the target therapy name, and \{\{example\}\} is replaced with in-domain Reddit/Twitter usage examples provided for stylistic guidance. [variant] was the placeholder for an LLM-generated variant.}
\label{fig:prompt_lex}
\end{figure*}

We adapted the QMisSpell misspelling generator \cite{SARKER201898}, which uses a phrase-level word embedding model trained on medication-related social media posts \cite{SARKER2017122}, to automatically generate common misspellings of medication names. To reduce noise, we restricted candidate variants to tokens that actually appeared in our Reddit TRD corpus and removed semantically similar but lexically distant neighbors via manual review. To further expand our lexicon, we employed generative LLM-based augmentation prompting a LLaMA~3-70B parameter model \cite{llama3modelcard} (prompt shown in \ref{fig:prompt_lex}.) For each therapy, we used a constrained few-shot prompt that (i) provided the generic name and (ii) supplied a few in-domain Reddit/Twitter examples illustrating how users misspell or refer to medications in free text (from QMisSpell and brand-name resources). We decoded with low-stochasticity sampling to prioritize precision over diversity (temperature = 0.2, nucleus sampling top-$p$ = 0.9; maximum 220 new tokens), and required the model to output only a plain list of candidate strings (one per line) with no explanations. Candidates were then merged with QMisSpell outputs and manually reviewed to remove implausible strings (e.g., non-medication words), ambiguous short forms, or terms that could map to multiple medications.

%For each seed alias in our initial dictionary, we retrieved up to 4{,}000 nearest neighbors from the word vector space, computed a Levenshtein-based similarity score against the seed term and retained only those candidates whose similarity exceeded a predefined threshold (0.70), following the weighted variant of QMisSpell that emphasizes early-character matches in order to reduce false positives \cite{SARKER201898}. To further suppress noise, we restricted candidate variants to tokens that actually appeared in our Reddit TRD corpus and removed semantically similar but lexically distant neighbors during manual review. 

The final lexicon contained 1{,}027 distinct lexical variants across 81 medications (median 12 variants per drug; range 6--19; Table~\ref{tab:med}). We applied this lexicon in a case-insensitive manner to each post’s title and body. To avoid false positives, we required that a match correspond to a full token or multiword expression rather than an arbitrary substring. Each detected surface form was then mapped back to its generic name. Multiple occurrences of the same medication within a post contributed multiple mention counts, while subscriber-level summaries counted each subscriber once per medication. After normalization, we obtained 23{,}399 medication mentions spanning 81 of the 83 medications; two medications, brexanolone and iloperidone, were not detected.

\subsection{Aspect-Based Sentiment Classifier Development}
We trained and employed an aspect-based sentiment classifier to label sentiment associated with each medication mention (negative/neutral/positive). Training data were obtained from the SMM4H 2023 English Twitter therapy-sentiment dataset \cite{10.1093/jamia/ocae010,GUO2023109618}, where each instance includes tweet text, a target therapy span, and a mention-level sentiment label. The original training set contained 3{,}009 tweets with significant class imbalance (319 negative, 2{,}120 neutral, 570 positive; Table~\ref{tab:twitter_absa_labels}). Here, counts refer to \emph{tweet-level instances} in the form of full tweets or posts: each instance contains one marked target-therapy span, and the sentiment label is assigned to that target mention within the tweet.

\begin{table}[ht]
\centering
\caption{Label distribution in the original and augmented Twitter therapy-sentiment training sets used for aspect-based sentiment analysis classification.}
\begin{tabular}{lrrrr}
\hline
Dataset & Negative & Neutral & Positive & Total \\
\hline
Original train & 319 (10.6\%) & 2{,}120 (70.4\%) & 570 (18.9\%) & 3{,}009 \\
Augmented train & 1{,}914 (25.7\%) & 2{,}120 (28.4\%) & 3{,}420 (45.9\%) & 7{,}454 \\
\hline
\end{tabular}
\label{tab:twitter_absa_labels}
\end{table}

\begin{figure*}[ht]
\includegraphics[width=\linewidth]{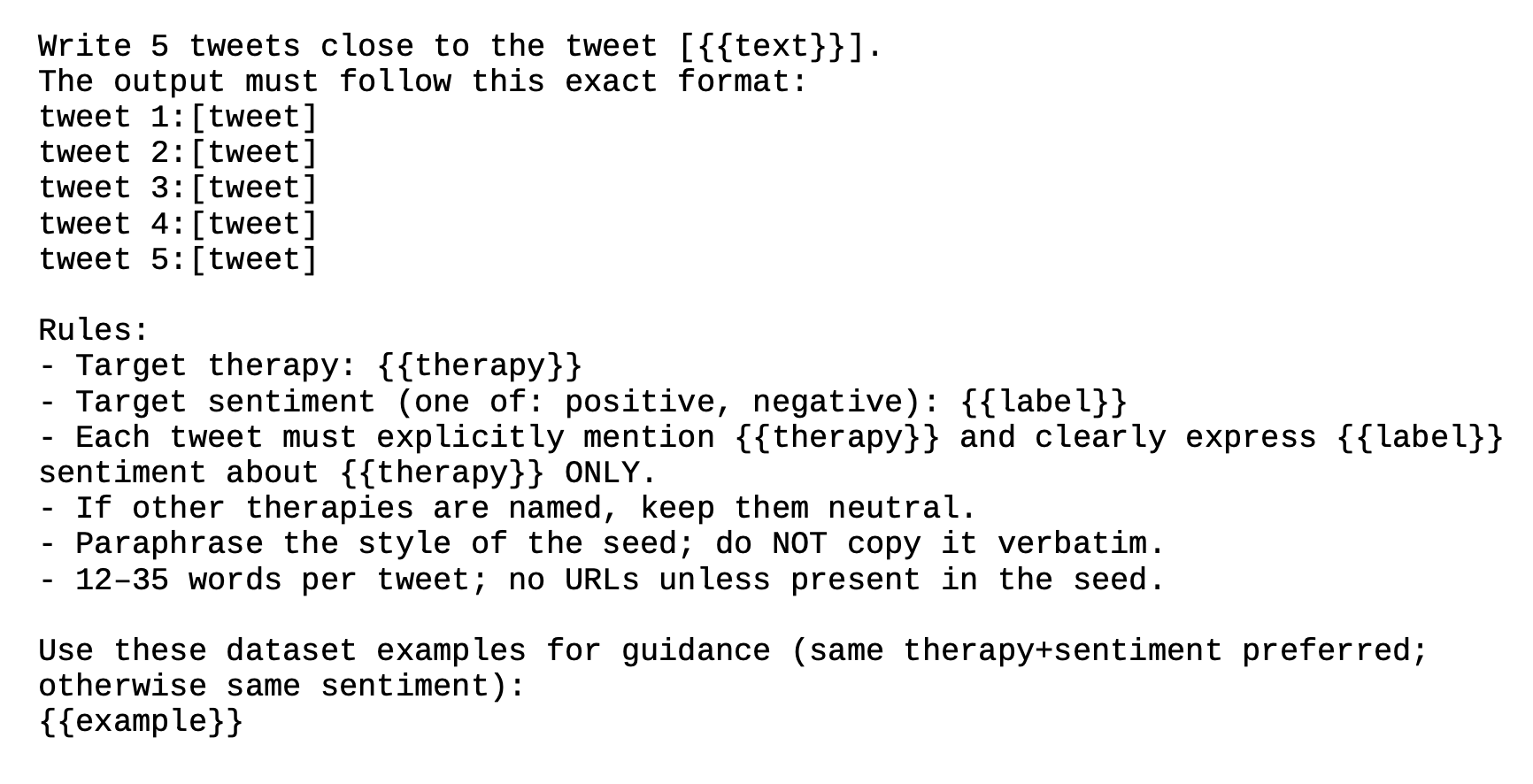}
\caption{Prompt used for LLM-based data augmentation. [text] was the placeholder for the original post, and [tweet] was the placeholder for an LLM-generated variant.}
\label{fig:prompt}
\end{figure*}

To address imbalance, we generated synthetic minority-class instances using LLaMA3-70B-Instruct \cite{llama3modelcard}. \textit{Original train} denotes the unmodified SMM4H 2023 training split, whereas \textit{Augmented train} denotes the same split after adding LLaMA3-generated synthetic tweets for the positive/negative classes. For each positive/negative post in the training set, we prompted the model (few-shot) to produce five synthetic posts that preserved (i) the same target therapy and (ii) the same sentiment toward that therapy, while keeping other therapies neutral (prompt shown in Figure~\ref{fig:prompt}). This yielded 4{,}445 synthetic tweets (2{,}850 positive; 1{,}595 negative), producing an augmented training set of 7{,}454 tweets; the test set was not altered.

We fine-tuned several Transformer models on this augmented dataset, including BERT-base \cite{devlin2019bert}, RoBERTa-base \cite{liu2019roberta}, and DeBERTa-v3-base \cite{he2021debertav3}, using a standard sequence-classification head with cross-entropy loss and early stopping based on validation performance. Before tokenization, the target therapy span in each tweet was replaced with a special placeholder token, \texttt{<MEDICATION>}. This tagging step forces the model to rely on the local linguistic context around the medication rather than on the specific surface form of the drug name, which is important for generalization to the diverse brand names, generics, and misspellings. For long texts, we further restricted attention to a window of up to 1{,}000 characters centered on the \texttt{<MEDICATION>} token. Classifier performance on the SMM4H test set was evaluated using micro-$F_1$ (primary metric) and macro-$F_1$, with class-specific $F_1$ scores. Uncertainty in micro-$F_1$ was quantified via nonparametric bootstrap (1{,}000 resamples) to obtain 95\% confidence intervals \cite{Efron1979}. The best-performing model, DeBERTa-v3, was used for classification of all medication-mentioning posts.

\subsection{Applying the Classifier on Reddit Data}
We extracted each medication-mentioning sentence along with its local context window (the sentence containing the mention with surrounding context) and post metadata (post ID, subreddit, subscriber, timestamp, sentence index), applying the same preprocessing as in the training experiments. We then used the fine-tuned DeBERTa-v3 classifier to predict a sentiment label for each mention. Predictions were mapped back to cohort metadata for analyses by medication, subscriber, subreddit, and year. We performed a manual spot-check on a sample of predictions to assess validity but did not relabel Reddit instances. The majority of sampled predictions were judged consistent with the post context. Table \ref{tab:sentiment_examples_by_therapy} collates illustrative, de-identified paraphrased excerpts per therapy, along with the assigned label and the model confidence $P(\hat{y})$, to facilitate qualitative comparison across therapies.

\subsection{Statistical Analysis}
We summarized the cohort at the post, medication, and subscriber levels. At the post level, we described temporal trends (posts per calendar year) and the normalized annual share for all TRD posts and the subset with $\geq$1 medication mention, computed as $\frac{\#\text{distinct posts in year }}{\#\text{distinct posts across all years in that cohort}}\times 100\%$. At the medication level, we computed (i) total mentions per medication and (ii) the number and proportion of subscribers mentioning each medication at least once. We also computed the number of distinct medications mentioned per subscriber. For each medication, we aggregated sentiment predictions across all mentions and reported the proportions of negative/neutral/positive labels as its sentiment profile. 

%Classifier performance on the SMM4H test set was evaluated using micro-averaged $F_{1}$ and macro-averaged $F_{1}$ over the three sentiment classes, along with class-specific $F_{1}$ scores for negative, neutral, and positive labels. Micro $F_{1}$, which in this single-label multiclass setting is equivalent to accuracy, served as the primary evaluation metric because it was the official shared task metric and reflects overall correctness weighted by class frequency. Macro $F_{1}$ provided a complementary view of performance that gives equal weight to each class regardless of prevalence. To quantify uncertainty, we used nonparametric bootstrap resampling to obtain $95\%$ confidence intervals for micro $F_{1}$. For each model, we generated $N = 1000$ bootstrap samples of the test set by sampling tweets with replacement, computed micro $F_{1}$ on each sample and took the $2.5^{\text{th}}$ and $97.5^{\text{th}}$ percentiles of the resulting distribution as the confidence interval bounds \cite{Efron1979}.

We conducted inferential analyses to assess whether sentiment differed (i) within individual medications and (ii) across medication categories. All tests used the DeBERTa medication-mention predictions on the Reddit TRD corpus ($N=23{,}399$ medication mentions).

\subsubsection{Therapy-level positive versus negative asymmetry.}
For each generic-name medication with at least one non-neutral mention, we tested whether positive and negative mentions occurred in equal proportion among non-neutral mentions (i.e., considering only Positive and Negative labels and excluding Neutral). Let $x$ be the number of Positive mentions and $n=x+y$ the number of non-neutral mentions (Positive $x$ plus Negative $y$) for a given medication. We applied an exact two-sided binomial test of $H_{0}: p=0.5$, where $p$ is the probability that a non-neutral mention is Positive. We report the estimated positive share $\hat{p}=x/n$, its 95\% confidence interval, and the Benjamini--Hochberg FDR-adjusted $p$-value across medications.

\subsubsection{Category-level differences in sentiment composition.}
To test whether sentiment composition differed by medication category, we constructed a $16\times 3$ contingency table crossing medication class (16 therapeutic classes) with sentiment label (Negative/Neutral/Positive) and performed a Pearson chi-square test of independence. We report $\chi^{2}$, degrees of freedom, and the $p$-value. Effect size was summarized using Cram\'er's $V$ \cite{cramer1999mathematical,mchugh2013chi}. To localize which class-sentiment cells contributed most strongly to the overall association, we examined standardized residuals from the fitted independence model. For post hoc class-pair comparisons, we conducted chi-square tests on all 120 $2\times 3$ class-pair subtables and controlled FDR across all class pairs using the Benjamini--Hochberg procedure.

\section{Results}

\subsection{Post and Subscriber Characteristics}

\begin{figure*}[ht]
\includegraphics[width=\linewidth]{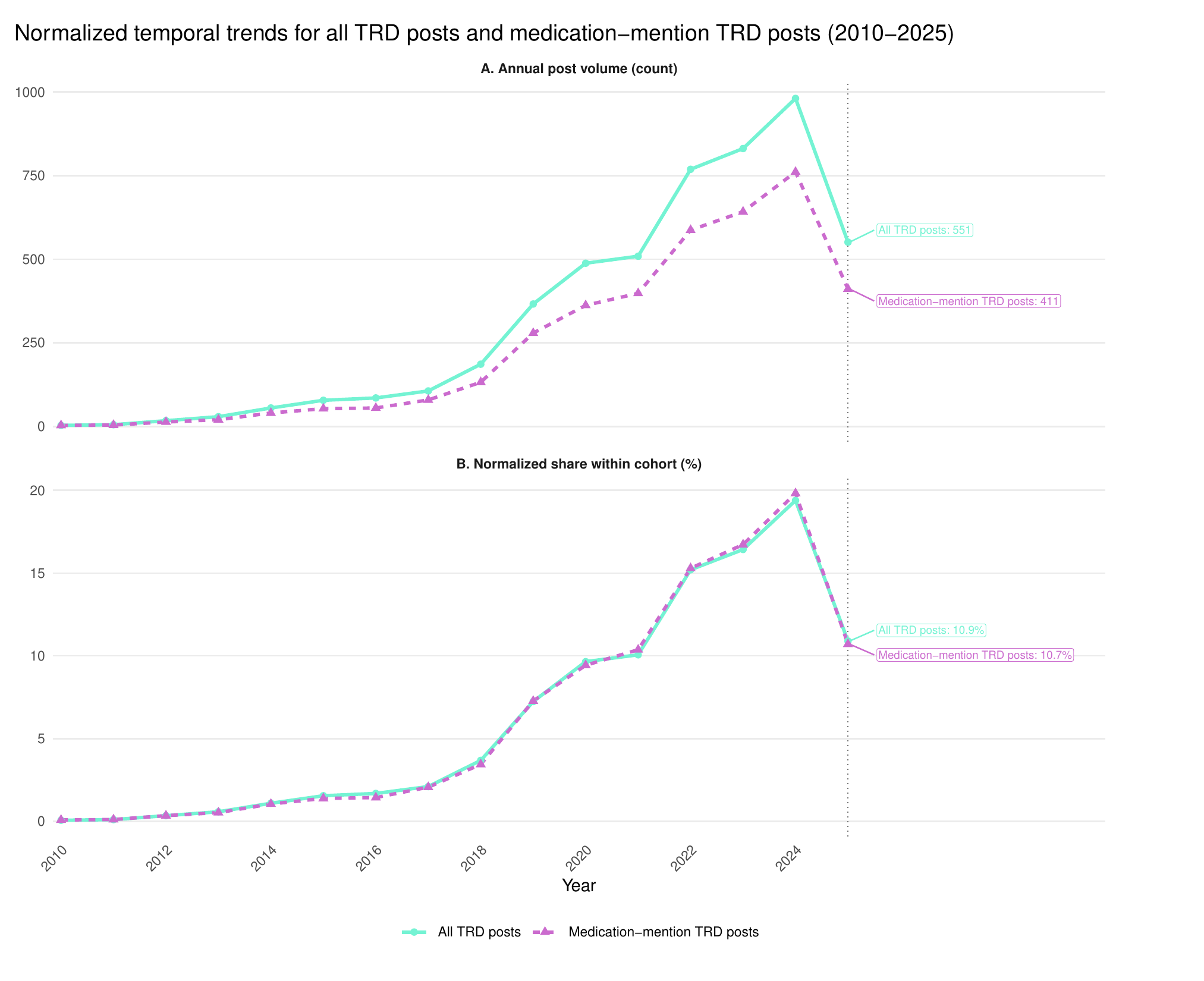}
\caption{Annual volume and normalized distribution of Reddit posts mentioning treatment-resistant depression and the medication-mention subset, 2010 to 2025. Two cohorts are shown: all TRD posts and TRD posts with $\geq$1 medication mention. \textbf{(A)} Annual post volume: lines plot the number of distinct posts per year for each cohort. \textbf{(B)} Normalized annual share within cohort: lines plot each year’s proportion of the cohort, computed as $\frac{\#\text{distinct posts in year }y}{\#\text{distinct posts across all years in that cohort}}\times 100\%$. The vertical dashed marker indicates that 2025 reflects a partial year through the end of data collection (July 2025).}
\label{fig:year}
\end{figure*}

\paragraph{Volume of TRD discussions over time.}
Figure~\ref{fig:year} summarizes the yearly volume of TRD posts (2010--July 2025) for both the full TRD cohort and the subset with $\geq$1 medication mention. Generally, TRD posts were rare before 2013 and increased substantially in later years. In the full cohort, annual post counts increased from 3--29 posts/year during 2010--2013 to 55 in 2014 and peaked at 981 in 2024; 551 posts were observed in 2025 through July. In the medication-mention subset, annual counts rose from 3--20 posts/year during 2010--2013 to 40 in 2014, 79 in 2017, 279 in 2019, 362 in 2020, 398 in 2021, and 761 in 2024; 411 posts were observed in 2025 through July.

\begin{figure}[H]
\includegraphics[width=\linewidth]{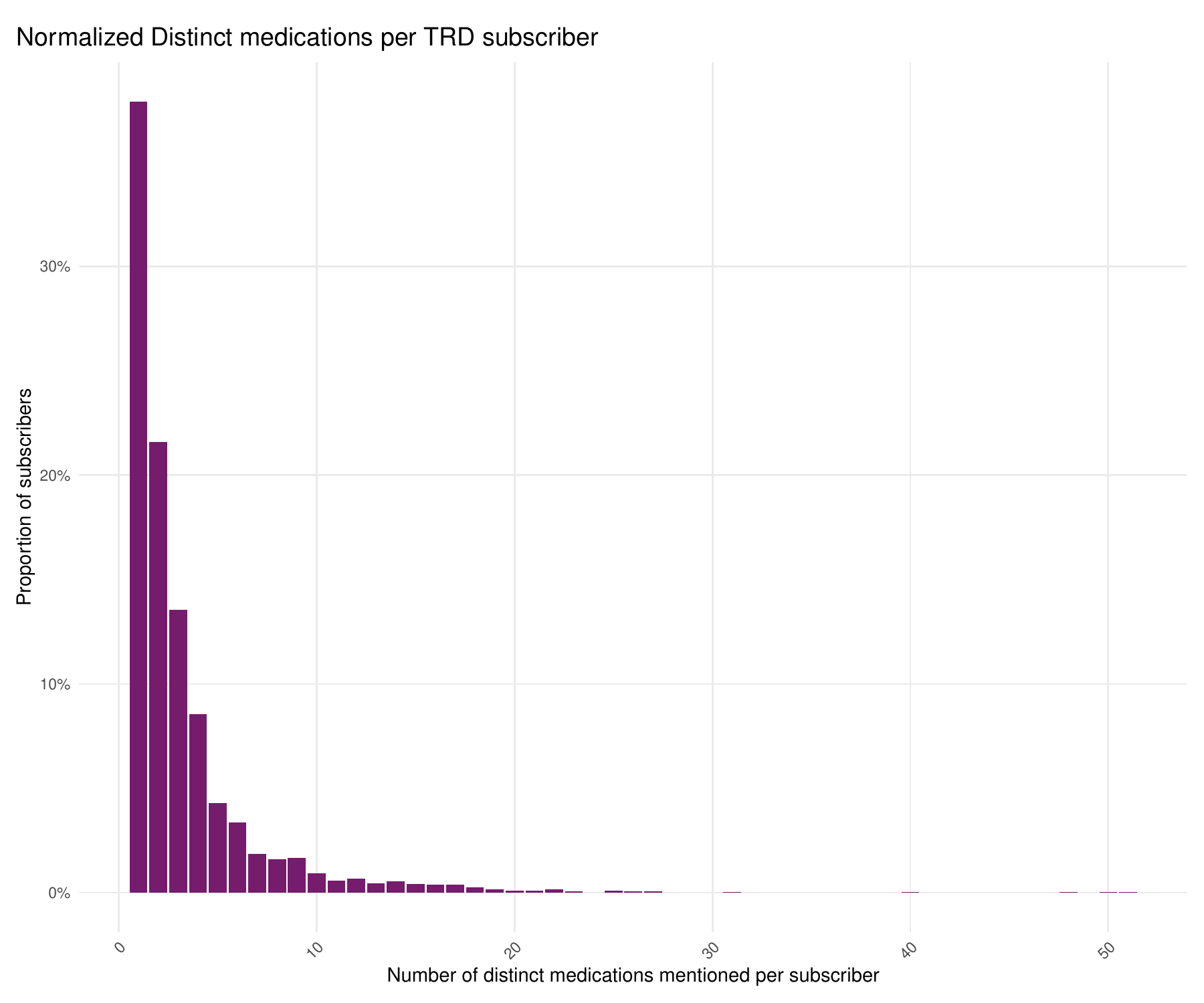}
\caption{Distribution of the number of distinct medications mentioned per subscriber in the TRD cohort. For each medication-mentioning subscriber ($N=2{,}700$), we counted the number of unique generic (nonproprietary) drug names referenced. The distribution is right-skewed, with most subscribers mentioning fewer than four medications.}
\label{fig:subscriber_hist}
\end{figure}

\paragraph{Subscriber engagement.}
Across all TRD posts, the dataset included 3{,}480 unique subscribers. Among subscribers who mentioned at least one medication ($N=2{,}700$), the number of distinct medications mentioned per subscriber was right-skewed (Figure~\ref{fig:subscriber_hist}): 37.9\% mentioned exactly one medication, 21.6\% mentioned two, and 13.6\% mentioned three; fewer than 5\% mentioned more than 10 medications. The maximum was 51 distinct medications for a single subscriber.

\subsection{Normalized Medication Mentions}

\begin{figure}[H]
\includegraphics[scale=0.3]{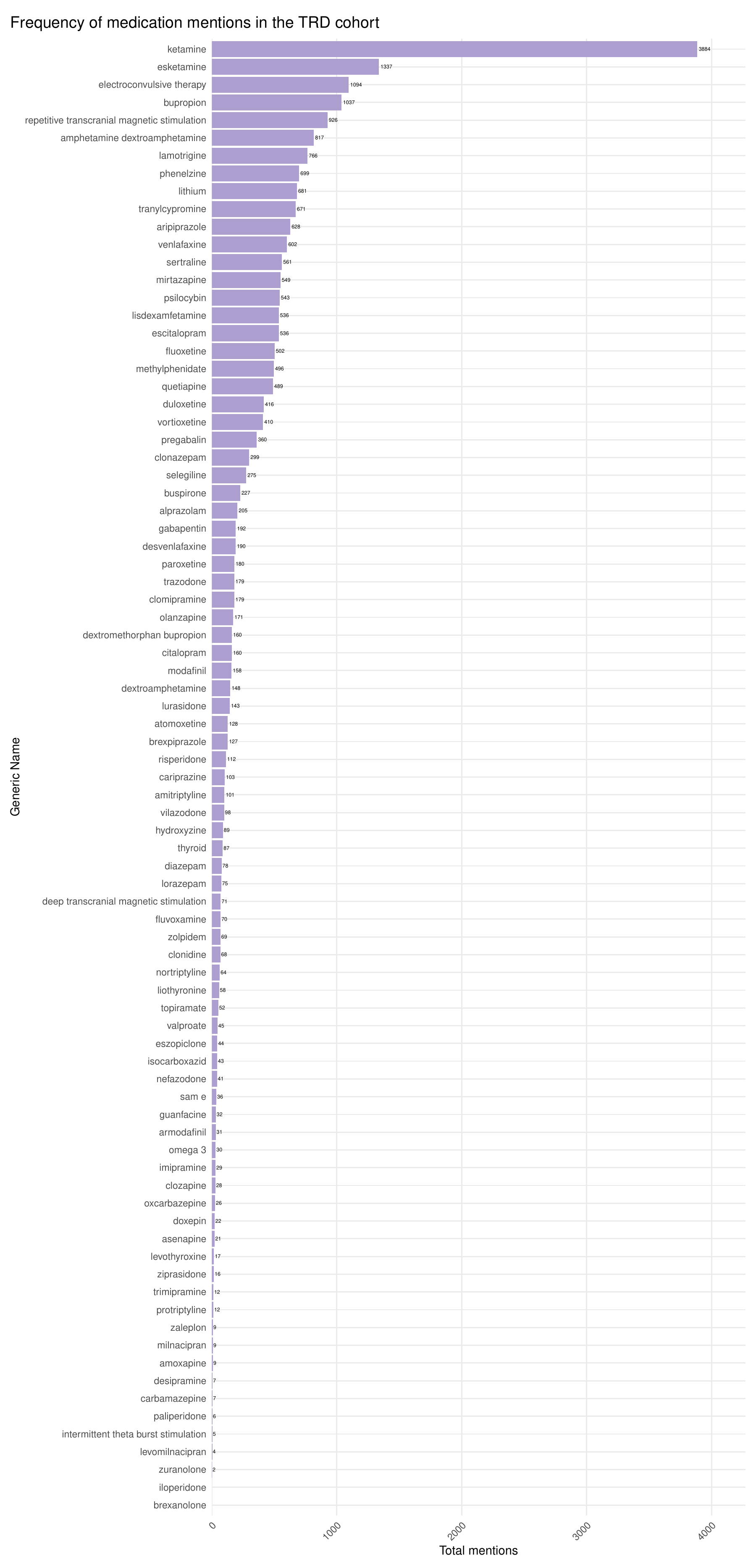}
\caption{Distribution of therapy-mention frequency in the Reddit TRD cohort. The plot shows total mention counts for each of 81 distinct nonproprietary therapy names that appeared at least once in the corpus, after collapsing spelling variants, brand names, and aliases to a single canonical form. Bars correspond to total mentions aggregated across all posts; therapies are ordered from most to least frequently mentioned. The head of the distribution is dominated by rapid-acting glutamatergic/NMDA-targeting therapies (e.g., ketamine and esketamine) and stimulant/ADHD-adjunct therapies (e.g., amphetamine--dextroamphetamine), whereas the remainder forms a long tail spanning diverse antidepressants, augmentation agents, neuromodulation procedures, and other adjunctive therapies.}
\label{fig:med_freq}
\end{figure}

We identified 23{,}399 total medication mentions from 81 medications that appeared at least once; brexanolone and iloperidone were not observed. The distribution of medication mention counts was highly skewed (Figure~\ref{fig:med_freq}). Ketamine was most frequent (3{,}884 mentions; 16.6\%), followed by esketamine (1{,}337; 5.7\%). Other high-frequency medications included electroconvulsive therapy (1{,}094), bupropion (1{,}037), repetitive transcranial magnetic stimulation (926), amphetamine-dextroamphetamine (817), lamotrigine (766), phenelzine (699), lithium (681), tranylcypromine (671), and aripiprazole (628). As an emerging therapeutic agent, psilocybin was mentioned 543 times (2.3\%).

\begin{figure}[ht]
\includegraphics[scale=0.3]{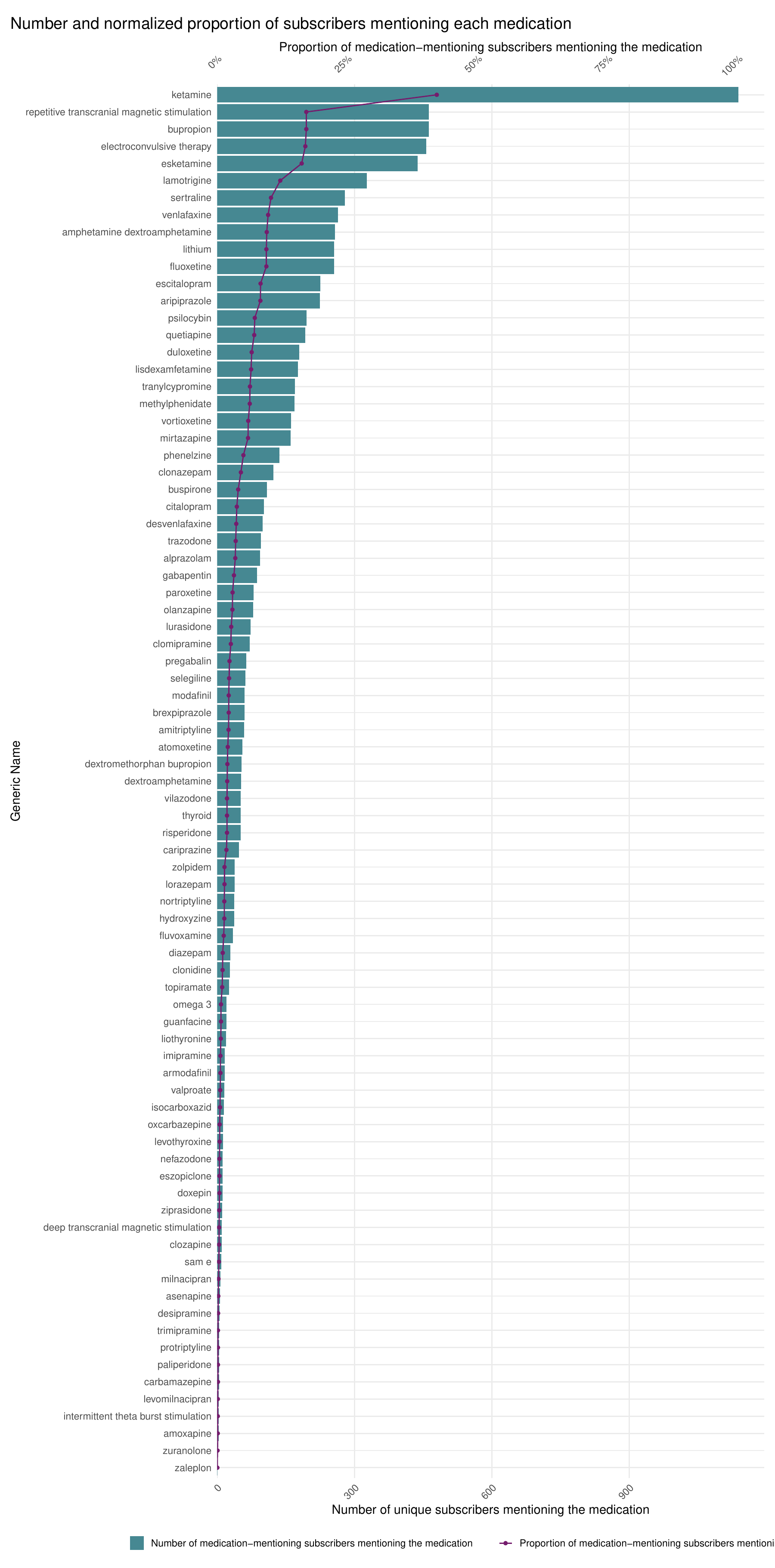}
\caption{Subscriber reach of each therapy in the Reddit TRD cohort. For each nonproprietary therapy name, we counted the number of \emph{unique} subscribers who mentioned it at least once. Horizontal bars show these subscriber counts ($u_i$). The overlaid line with points (top x-axis) shows the corresponding normalized reach among therapy-mentioning subscribers, $p_i = u_i/U$, where $U$ is the number of subscribers who mentioned at least one therapy in the cohort. Therapies are ordered from highest to lowest $u_i$.}
\label{fig:subscriber_drug}
\end{figure}

To quantify how widely each medication was discussed, we counted unique subscribers mentioning each medication at least once (Figure~\ref{fig:subscriber_drug}). Ketamine was mentioned by approximately 1{,}138 unique subscribers (\textasciitilde42\% of therapy-mentioning subscribers). Bupropion and repetitive transcranial magnetic stimulation were each mentioned by about 462 subscribers; electroconvulsive therapy by about 457; and esketamine by around 438. Some medications had relatively high mention counts concentrated among fewer subscribers (e.g., pregabalin: 360 mentions from 64 subscribers; selegiline: 275 mentions from 62 subscribers; alprazolam: 205 mentions from 94 subscribers).

\subsection{Aspect-based Sentiment Analysis of Drugs}

\subsubsection{Classifier selection}
As shown in Table~\ref{tab:absa_performance}, all three Transformer backbones achieved competitive performance on the SMM4H 2023 Task~2 test set, and LLM-based data augmentation yielded small but consistent gains in both micro and macro $F_{1}$ scores. Among the augmented models, DeBERTa-v3-base performed best, with a micro-$F_{1}$ of 0.800 (95\% CI 0.780–0.820), exceeding the top system reported in the shared task overview (0.778) \cite{10.1093/jamia/ocae010}. Its class-specific $F_{1}$ scores were well balanced across negative (0.467), neutral (0.874), and positive (0.636) sentiment. We therefore adopted the augmented DeBERTa-v3-base model for analyses.

\begin{table}[ht]
\centering
\caption{Performance of transformer-based classifiers on the SMM4H 2023 Task~2 test set (N = 1{,}602 tweets) before (baseline) and after LLM data augmentation (augmented). Micro-F$_1$ equals accuracy. Macro-F$_1$ is the unweighted mean of class-specific F$_1$ scores.}
\begin{tabular}{lccccc}
\hline
Model & Micro-$F_{1}$ (95\% CI)  & Macro-F$_{1}$ & $F_{1}$ (negative) & $F_{1}$ (neutral) & $F_{1}$ (positive) \\
\hline
BERT-base (baseline)          & 0.758 (0.738, 0.780) &  0.544 & 0.225 & 0.858 & 0.549 \\
RoBERTa-base (baseline)       & 0.782 (0.762, 0.802)
 & 0.633 & 0.416 & 0.871 & 0.612 \\
DeBERTa-v3-base (baseline)     & 0.780 (0.760, 0.801) & 0.656 & 0.466 & 0.869 & 0.633 \\
BERT-base (augmented)           & 0.768 (0.748, 0.789) & 0.619 & 0.420 & 0.860 & 0.577 \\
RoBERTa-base (augmented)      & 0.785 (0.765, 0.805) & 0.639 & 0.461 & 0.869 & 0.587 \\
DeBERTa-v3-base (augmented)   & 0.800 (0.780, 0.820) & 0.659 & 0.467 & 0.874 & 0.636 \\
\hline
\end{tabular}
\label{tab:absa_performance}
\end{table}

\subsubsection{Medication-specific sentiment profiles}
\begin{figure}[ht]
\includegraphics[scale=0.35]{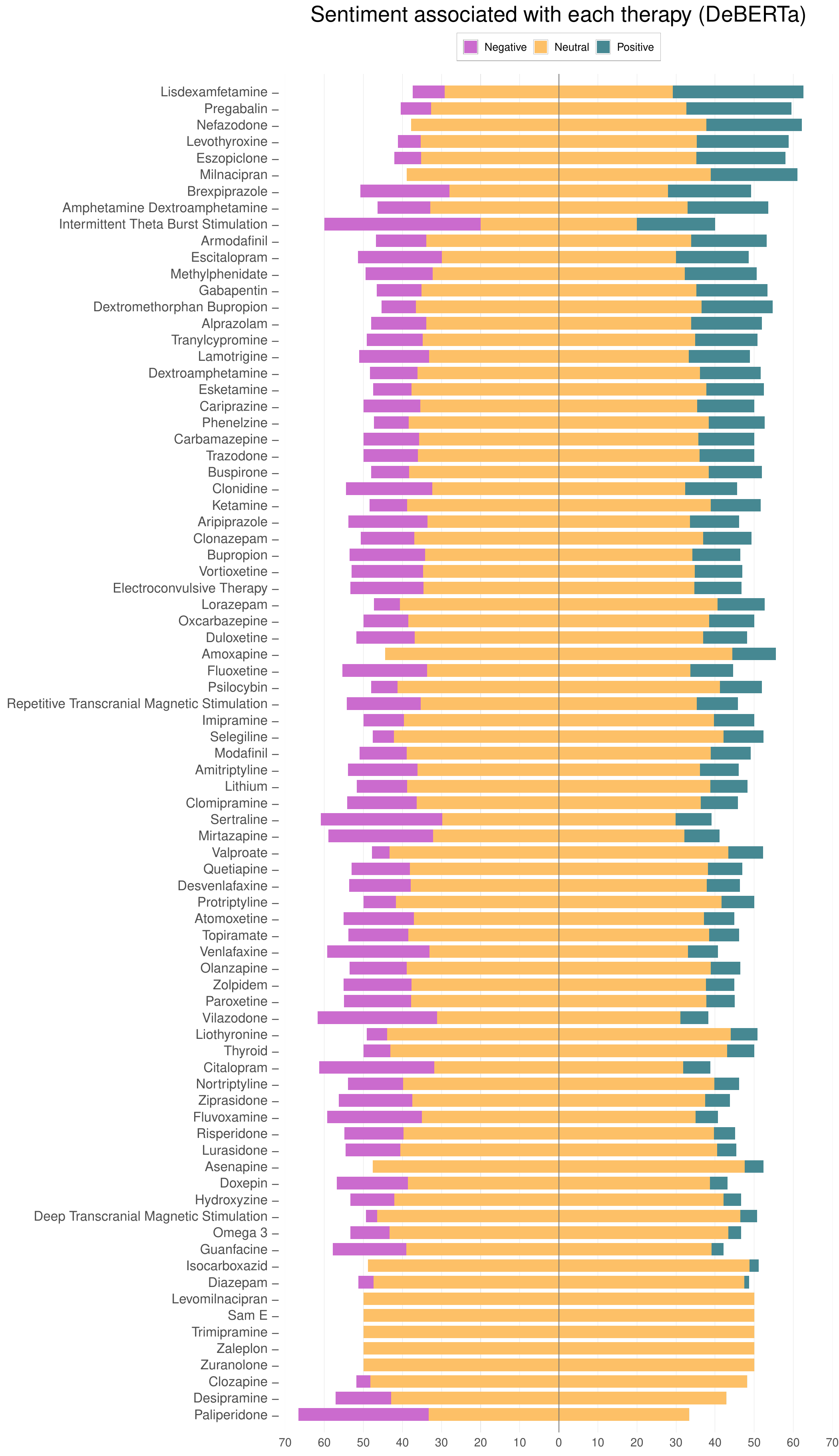}
\caption{\textbf{Therapy-specific sentiment in TRD-focused Reddit discourse.}
For each normalized medication, the horizontal stacked barshows the proportion of mentions labeled \emph{negative}, \emph{neutral}, or \emph{positive}, summing to 100\%. Negative proportions extend left, positive right, with neutral centered. Therapies are ordered by decreasing positive proportion. }
\label{fig:sentiment}
\end{figure}

Figure~\ref{fig:sentiment} shows mention-level sentiment proportions (negative/neutral/positive) for 81 medications using the DeBERTa model.

\subsubsection{Statistical Tests of Sentiment Differences}
\paragraph{Sentiment composition differs significantly across medication classes.}
Across all medication mentions ($N=23{,}399$), 16{,}865 (72.1\%) were labeled Neutral, 3{,}460 (14.8\%) Negative, and 3{,}074 (13.1\%) Positive. Sentiment composition differed significantly by medication class (Pearson chi-square test: $\chi^{2}(30,\,N=23{,}399)=686.07$, $p=3.93\times 10^{-125}$; Cram\'er's $V=0.121$), indicating modest but reliable shifts in sentiment proportions across categories rather than a uniform mixture of positive, neutral, and negative evaluations.

\FloatBarrier
\begin{figure}[H]
\centering
\includegraphics[width=\linewidth]{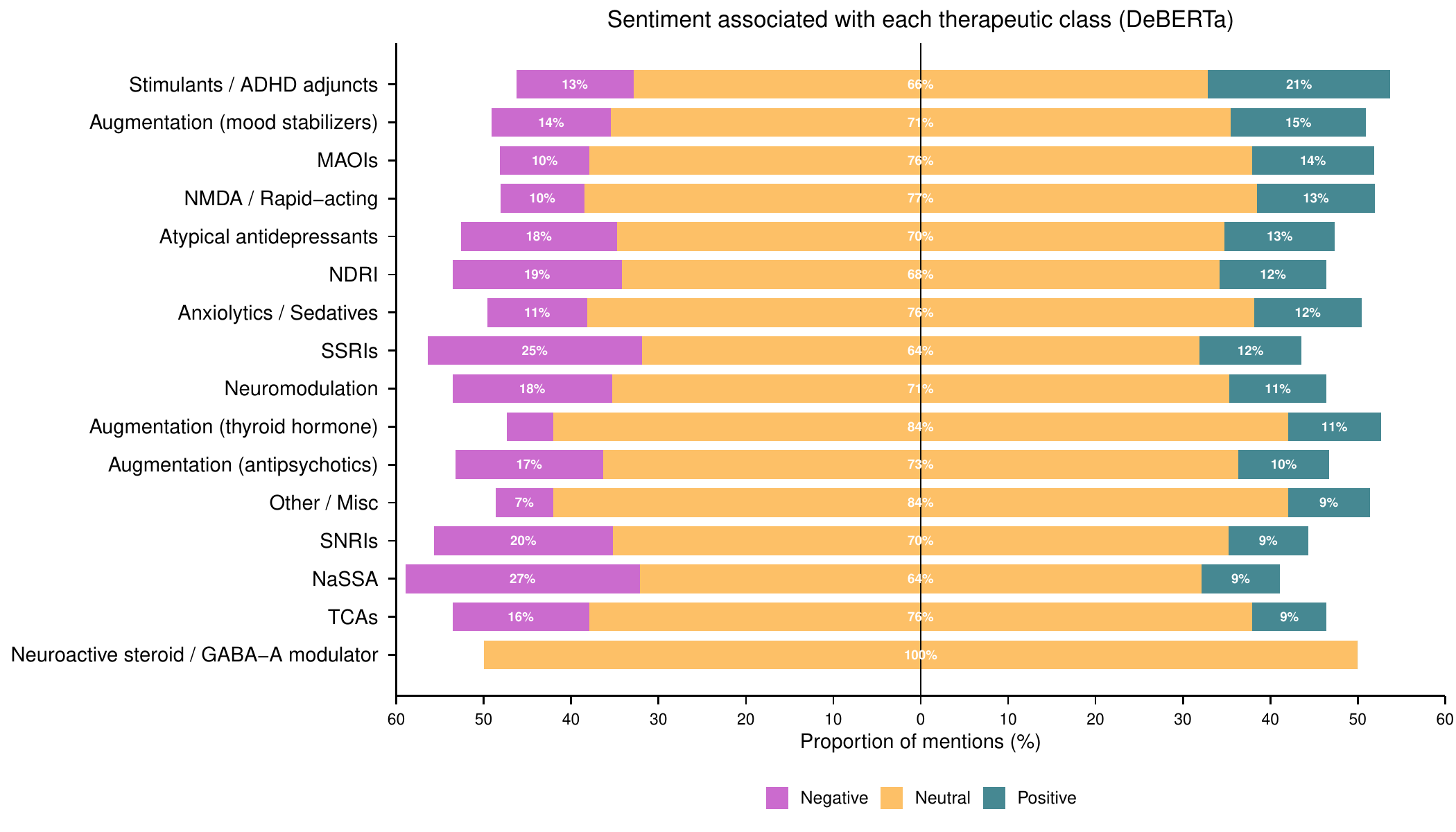}
\caption{Sentiment composition of therapy mentions by therapeutic class in TRD-focused Reddit discourse. Each bar represents the proportion of Reddit mentions within a therapeutic class labeled negative (red, extending left), neutral (orange, centered), or positive (green, extending right). Classes are ordered by descending positive share (top to bottom). }
\label{fig:class_sentiment_composition}
\end{figure}
\FloatBarrier

As shown in Figure~\ref{fig:class_sentiment_composition}, class-wise proportions highlighted distinct category signatures. The highest-volume classes were NMDA/rapid-acting (5{,}381), stimulants/ADHD adjuncts (2{,}414), augmentation (mood stabilizers) (2{,}129), neuromodulation (2{,}096), and SSRIs (2{,}009). NMDA/rapid-acting therapies retained a comparatively low negative share (515/5{,}381; 9.6\%) with a large neutral component (4{,}144/5{,}381; 77.0\%). In contrast, SSRIs (493/2{,}009; 24.5\% negative) and SNRIs (250/1{,}221; 20.5\% negative) showed higher negative proportions, and NaSSA had the highest negative proportion among classes with substantial counts (147/549; 26.8\%). Stimulants/ADHD adjuncts showed the highest positive proportion (504/2{,}414; 20.9\%). In post hoc analyses, many class pairs differed significantly after FDR correction (83/120 pairs), with particularly strong contrasts involving NMDA/rapid-acting therapies versus multiple antidepressant classes (e.g., NMDA/rapid-acting vs SSRIs, $p_{\mathrm{FDR}}=3.77\times 10^{-59}$; NMDA/rapid-acting vs NaSSA, $p_{\mathrm{FDR}}=1.28\times 10^{-31}$; NMDA/rapid-acting vs SNRIs, $p_{\mathrm{FDR}}=5.82\times 10^{-26}$).

\paragraph{Therapy-level tests identify drugs with asymmetric positive versus negative evaluations.}
Among 83 generic-name medications in the corpus, 76 had at least one non-neutral mention and were eligible for therapy-level testing. After FDR correction, 20/76 medications showed a statistically significant departure from parity between Positive and Negative mentions among non-neutral evaluations, as shown in Table \ref{tab:medication_pfdr}.

\begin{table*}[ht]
\centering
\small
\caption{Selected significant therapy-level binomial tests from classifier predictions.}
\begin{tabular}{p{0.31\textwidth}rrccc}
\hline
Medication & Positive $n$ & Negative $n$ & $\hat{p}$ (Positive) & 95\% CI for $\hat{p}$ & $p_{\mathrm{FDR}}$ \\
\hline
Nefazodone & 10 & 0 & 1.000 & [0.692, 1.000] & $9.28\times 10^{-3}$ \\
Lisdexamfetamine & 179 & 44 & 0.803 & [0.744, 0.853] & $1.25\times 10^{-18}$ \\
Pregabalin & 97 & 28 & 0.776 & [0.693, 0.846] & $6.38\times 10^{-9}$ \\
Phenelzine & 100 & 62 & 0.617 & [0.538, 0.692] & $1.57\times 10^{-2}$ \\
Vilazodone & 7 & 30 & 0.189 & [0.080, 0.352] & $1.21\times 10^{-3}$ \\
Citalopram & 11 & 47 & 0.190 & [0.099, 0.314] & $2.57\times 10^{-5}$ \\
Fluvoxamine & 4 & 17 & 0.190 & [0.054, 0.419] & $2.88\times 10^{-2}$ \\
Venlafaxine & 46 & 158 & 0.225 & [0.170, 0.289] & $3.59\times 10^{-14}$ \\
\hline
\end{tabular}
\label{tab:medication_pfdr}
\end{table*}

Positively skewed medications spanned multiple therapeutic classes, including \textit{Atypical antidepressants} (nefazodone), \textit{Stimulants / ADHD adjuncts} (lisdexamfetamine), \textit{Augmentation (mood stabilizers)} (pregabalin), and \textit{MAOIs} (phenelzine). In contrast, negatively skewed medications were concentrated in first-line antidepressant classes, particularly \textit{SSRIs} (citalopram, fluvoxamine) and an \textit{SNRI} agent (venlafaxine), alongside an \textit{Atypical antidepressant} agent (vilazodone). Collectively, these therapy-level tests quantify therapy-specific deviations from balanced evaluative sentiment and complement the broader class-level heterogeneity observed in the contingency analyses.

\section{Discussion}

\subsection{Principal Findings}
In this NLP-driven observational study, we characterized medication-related chatter from Reddit in terms of sentiment by combining (i) lexicon-based medication normalization and (ii) aspect-based sentiment classification. Three empirical findings stand out. First, TRD-relevant posting volume increased markedly over the study period (Figure~\ref{fig:year}). Second, medication discussion was concentrated in a subset of treatments: ketamine and esketamine were the most frequently mentioned medications and were referenced by large fractions of the subscriber base (Figures~\ref{fig:med_freq}--\ref{fig:subscriber_drug}). Third, sentiment was predominantly neutral overall, but sentiment composition differed significantly across medication classes and across individual medications (Figure~\ref{fig:sentiment}). These findings suggest growing online attention to TRD, potentially driven by greater awareness, expanding treatment options, and increased social media engagement around mental health, and present possible hypotheses for further exploration of pharmaceutical treatments in clinical settings. Our study also demonstrates the utility of social networking platforms such as Reddit for investigating patient perceptions of pharmaceutical treatments at scale, particularly for topics with limited data from clinical settings. 

\subsection{Increasing Trend in TRD Discourse}
The sharp rise in TRD-related posting after the mid-2010s likely reflects a combination of increased public and clinical attention to TRD and expanding awareness of and access to newer treatment options. Notably, the timing of the steep increase in the early 2020s coincides with broader growth of online mental health discussion during the COVID-19 period, which has been documented in Reddit mental health communities \cite{low2020natural,biester2021understanding}. In addition, esketamine’s FDA approval in March 2019 introduced a novel TRD therapy, and may have contributed to heightened discussion of rapid-acting interventions \cite{fda2019_esketamine_approval}. While these interpretations are based on temporal patterns, they do not represent causal conclusions, because the present study is observational.

\subsection{Therapy Frequencies}
Ketamine and esketamine, rapid-acting glutamatergic NMDA-receptor antagonists, dominated both mention counts and unique-subscriber reach, suggesting that rapid-acting NMDA-pathway therapies occupy a central place in contemporary TRD peer discussions. The high frequencies align with the clinical prominence of esketamine trials and approvals for TRD populations \cite{popova2019efficacy}. At the same time, intranasal esketamine is FDA-approved not only for TRD but also for depressive symptoms in adults with major depressive disorder with acute suicidal ideation or behavior; accordingly, a subset of esketamine-related discussion may reflect indication overlap rather than TRD specifically \cite{spravato_fda_label}. Conventional antidepressants (e.g., SSRIs/SNRIs) and common augmentation agents also remained widely discussed, which is expected in a TRD cohort because patients typically report multiple prior trials and often continue antidepressants while also being offered interventional psychiatric strategies.

In addition, neuromodulation therapies (e.g., rTMS and ECT) showed high subscriber reach, demonstrating broad community-level salience. In contrast, stimulant/ADHD-adjunct therapies (e.g., amphetamine--dextroamphetamine, lisdexamfetamine, methylphenidate, and atomoxetine) were among the most frequently referenced adjunctive categories by mention volume, while overall discussion extended into a long tail spanning diverse antidepressants, augmentation strategies, and other adjunctive therapies.

Subscriber-level breadth showed a strongly right-skewed pattern: most subscribers referenced only one to a few medications, whereas a small minority described extensive medication histories. Such long-tailed dispersion is expected in TRD management that typically involves stepwise switching and/or augmentation following partial or nonresponse, producing wide heterogeneity in cumulative medication exposure across patients \cite{doi:10.1176/ajp.2006.163.11.1905}. Medications such as pregabalin or selegiline had substantial mention counts concentrated among relatively few subscribers, which may reflect a smaller set of subscribers posting repeated updates about a specific regimen (e.g., dosage, side effects, or trial sequences). In contrast, medications with both high mention counts and high subscriber reach are more likely reflect ``community-wide'' topics rather the activity of a small number of highly active posters. 

%Mention volume quantify discussion intensity (how often a medication is referenced). Reach reflects diffusion across the cohort (how many unique subscribers reference it at least once). Accordingly, a high-mentions yet low-reach profile suggests concentrated, iterative reporting by a small subset of subscribers. High reach, by contrast, indicates broad, shared relevance even when per-subscriber discussion is brief.

We therefore distinguish \emph{discussion intensity} (how often a medication is referenced) from \emph{diffusion} (how many unique subscribers reference it at least once). A high-mentions but low-reach profile suggests concentrated, iterative reporting by a small subset of subscribers, whereas high reach indicates broad shared relevance even when per-subscriber discussion is brief.

\subsection{Therapy Sentiment Profiles}
Across all mentions, over 70\% were neutral, which is typical of health-related social media where users frequently list treatments, doses, or histories without an explicit sentiment. Consequently, sentiment proportions should not be interpreted as clinical response rates. Instead, they capture how often TRD posters express explicitly positive or negative evaluations when discussing a medication. Despite the high prevalence of neutral sentiment, significant variations were observed by medication class. NMDA/rapid-acting therapies had a lower negative share than several conventional antidepressant classes, whereas SSRIs/SNRIs and NaSSA classes showed higher negative proportions. These differences are compatible with the clinical context that defines TRD: many users have already experienced inadequate response to standard antidepressants and may therefore reference SSRIs/SNRIs primarily in the context of prior nonresponse with accompanying negative sentiment. Conversely, rapid-acting therapies may be discussed as later-line or novel options, and randomized withdrawal data for intranasal esketamine in TRD suggest that symptom gains are typically maintained only with ongoing maintenance dosing, with continuation significantly reducing relapse risk among responders \cite{daly2019efficacy}.

At the medication level, binomial tests among non-neutral mentions further support the class-level patterns (see Table \ref{tab:full_medication_pfdr}.) Ketamine and esketamine showed modest but significant positive skew (ketamine: 496 positive vs 370 negative; esketamine: 197 vs 131), whereas several traditional antidepressants were strongly negatively skewed (e.g., sertraline: 52 vs 174; venlafaxine: 46 vs 158). Augmentation agents such as aripiprazole (79 vs 127) and quetiapine (43 vs 73) were also negatively skewed among non-neutral mentions. These findings do not necessarily verify clinical efficacy. Rather, they indicate that, within the TRD Reddit cohort, sentiments associated with certain medications are disproportionately unfavorable relative to others.

Stimulants/ADHD adjuncts exhibited the highest positive proportion at the class level and included positively skewed medications (e.g., lisdexamfetamine). This pattern likely reflects multiple mechanisms that are not separable in the current data, such as off-label use, comorbid ADHD treatment, or patient emphasis on energy and motivation. Clinical trial evidence for stimulant augmentation in depression is mixed, so these signals should be considered hypothesis-generating and interpreted in light of selection and reporting biases in social media \cite{richards2016lisdexamfetamine,patkar2006randomized}.

\subsection{Methodological Implications}
Medication normalization is necessary for valid corpus-level comparisons in noisy social media text. Without consolidating brand names, abbreviations, and misspellings into generic names, high-frequency medications would fragment into multiple strings, distorting both frequency and sentiment summaries. Furthermore, mention-level (target-dependent) sentiment is crucial in this setting because posts frequently contain multiple medications with different sentiment polarities. Thus, assigning a single post-level polarity would conflate targets. Finally, the augmentation strategy involving LLM-generated synthetic data improved classification performance, consistent with recent research \cite{Guo2024LLMHealthTextClassification}.

\subsection{Limitations}
The findings reported in this paper should be interpreted with several limitations in mind. Reddit subscribers may not be representative of the broader TRD population, and those with persistent or complex experiences may be more likely to post. Therefore, medication and sentiment distributions should not be interpreted as population estimates. Social media users tend to be younger than the general patient population. All information is self-reported. Thus, TRD status, adequacy of prior trials, diagnoses, dosing, adherence, and comorbidities cannot be verified. The analysis is also restricted to posts containing explicit medication names covered by the lexicon; implicit treatment references, non-pharmacologic therapies, and self-management strategies are likely under-captured. In terms of the NLP component, the sentiment classifier may mishandle sarcasm, vague sentiments, or highly context-dependent statements. We did not build a human-annotated Reddit test set to quantify domain-specific error, so there may be undetected errors caused by domain shift between social networks. Additionally, the classifier's negative-class $F_{1}$ score (0.467) was notably lower than for other classes, which may lead to underestimation of negative sentiment proportions. Finally, observational aggregation cannot separate perceived efficacy from tolerability, access barriers, or life events; nor can it infer causal effects of therapies.

\section{Conclusion}

Our analysis shows that TRD-focused peer-support discourse on Reddit can be systematically  converted into structured, therapy-level sentiment profiles by combining lexicon-based medication normalization with mention-level, aspect-based sentiment analysis. To our knowledge, this represents the first large-scale quantitative characterization of TRD medication discussions on Reddit, leveraging a multi-community cohort and a standardized lexicon to consolidate brand names, colloquialisms, and misspellings into generic drug representations. The resulting pipeline is reusable and scalable for comparing patient-reported treatment evaluations in settings where labeled data are limited and medication mentions are highly variable.

Future work should strengthen clinical interpretability by further characterizing non-neutral sentiments into specific drivers, such as efficacy, adverse effects, discontinuation and loss of effect, and by extracting clinically relevant context such as dose, duration, and co-therapy patterns. Methodologically, a modest human-annotated Reddit evaluation set would enable explicit domain calibration, and longitudinal subscriber-level analyses could characterize within-person changes around treatment initiation and switching while maintaining privacy-preserving aggregation. Extending the framework to neuromodulation and psychotherapy would broaden coverage of TRD management beyond pharmacotherapy. Finally, combining social media data with clinical data sources, such as electronic health records, may provide more in-depth insights through the integration of complementary patient- and clinician-generated knowledge.

\clearpage

%Bibliography
\bibliographystyle{unsrt}  
\bibliography{references}

\clearpage

\section*{Supplementary Material}
\beginsupplement

%\begin{landscape}
\small
\setlength{\tabcolsep}{3pt}
\renewcommand{\arraystretch}{1.1}

\begin{longtable}{p{0.25\textwidth} p{0.25\textwidth} p{0.5\textwidth}}
\caption{\textbf{The full list of medications included in the Reddit TRD corpus and their misspellings/lexical variants.}}\label{tab:med}\\
\hline
\textbf{Class} & \textbf{Medication} & \textbf{Misspellings and Variants} \\
\hline
\endfirsthead

\hline
\textbf{Class} & \textbf{Medication} & \textbf{Misspellings and Variants} \\
\hline
\endhead

\hline
\multicolumn{3}{r}{\textit{Continued on next page}}\\
\hline
\endfoot

\hline
\endlastfoot

Anxiolytics / Sedatives & alprazolam & alprazlam, alprazolm, alprzolam, alparzolam, alpraozlam, alprazalam, alprazloam, alprazoalm, xanax bars, zannies, zanz, bars, xanax, xanex \\ \hline
Anxiolytics / Sedatives & clonazepam & clnazepam, clonazepm, clonazpam, clonzepam, clanazepam, clnoazepam, cloanzepam, clonaezpam, klonies, klons, clonazepam, klonopin, bromazepam, klonipin, clonazapam, clonopin \\ \hline
Anxiolytics / Sedatives & diazepam & dazepam, diazepm, diazpam, dizepam, daizepam, deazepam, diaezpam, diazeapm, vallies, diazepam, diazapam, aluminum \\ \hline
Anxiolytics / Sedatives & lorazepam & lorazepm, lorazpam, lorzepam, lrazepam, larazepam, loarzepam, loraezpam, lorazeapm, lorazepam, atavan \\ \hline
Anxiolytics / Sedatives & buspirone & buspiron, buspar, bspirone, buspirne, busprone, bospirone, bsupirone, bupsirone, busiprone, busperone, Buspar, buscopan \\ \hline
Anxiolytics / Sedatives & hydroxyzine & hydroxyzene, hydroxazine, hydroxyzin, hydroxyzne, hydrxyzine, hdyroxyzine, hidroxizine, hydorxyzine, hydraxyzine, hydroxyizne, Vistaril, Atarax, hydroxyzine \\ \hline
Anxiolytics / Sedatives & zolpidem & zlpidem, zolpdem, zolpidm, ozlpidem, zalpidem, zlopidem, zolipdem, zolpdiem, ambien \\ \hline
Anxiolytics / Sedatives & eszopiclone & eszopclone, eszopiclne, eszpiclone, esozpiclone, eszapiclane, eszoipclone, eszopcilone, eszopeclone, zopiclone, eszopiclone, zopliclone \\ \hline
Anxiolytics / Sedatives & zaleplon & zalepln, zalplon, zleplon, azleplon, zaelplon, zalelpon, zaleplan, zaleplno \\ \hline
Atypical antidepressants & vortioxetine & vortioxitine, vortioksetine, vortioxetin, vortioxetne, vortioxtine, vortixetine, vortoxetine, vrtioxetine, ovrtioxetine, vartiaxetine, Trintellix, Brintellix \\ \hline
Atypical antidepressants & trazodone & trazadone, trazodonee, trazdone, trazodne, trzodone, rtazodone, tarzodone, traozdone, trazadane, trazdoone, Desyrel, Oleptro, trazodone, trazedone \\ \hline
Atypical antidepressants & vilazodone & vilazidone, vilazodine, vilazdone, vilazodne, vilzodone, vlazodone, ivlazodone, velazodone, vialzodone, vilaozdone, Viibryd \\ \hline
Atypical antidepressants & nefazodone & nefazdone, nefazodne, nefzodone, nfazodone, enfazodone, neafzodone, nefaozdone, nefazadane \\ \hline
Augmentation (antipsychotics) & aripiprazole & aripriprazole, aripiprazol, abilify, aripiprazle, aripiprzole, aripprazole, arpiprazole, airpiprazole, arepeprazole, ariipprazole, aripiparzole, Abilify, Abilify Maintena, Abilify MyCite, aripiprozole \\ \hline
Augmentation (antipsychotics) & quetiapine & quetapine, quetiapene, quetiapinee, qetiapine, quetiapin, quetiapne, quetipine, qutiapine, qeutiapine, qoetiapine, Seroquel, Seroquel XR, quell, susie q, baby heroin, seroquil, seraquel, seraquil \\ \hline
Augmentation (antipsychotics) & olanzapine & olanzapene, olazapine, olanzapin, olanzapne, olanzpine, olnzapine, alanzapine, loanzapine, oalnzapine, olanazpine, Zyprexa, Zyprexa Zydis, Symbyax, olanzapine, olanzipine \\ \hline
Augmentation (antipsychotics) & risperidone & risperdal, risperadone, risperidonee, risperdone, risperidne, rispridone, rsperidone, irsperidone, resperedone, ripseridone, risepridone, Risperdal, Risperdal Consta, risperidone, risperidal \\ \hline
Augmentation (antipsychotics) & asenapine & asenapin, asenapne, asenpine, asnapine, aesnapine, aseanpine, asenaipne, asenapene, Saphris, Secuado, asenapine \\ \hline
Augmentation (antipsychotics) & lurasidone & lurasidonee, lurasadone, lrasidone, lurasdone, lurasidne, lursidone, lorasidone, lruasidone, luarsidone, luraisdone, Latuda, lurasidone \\ \hline
Augmentation (antipsychotics) & brexpiprazole & brexiprazole, brexiprazol, brexpiprazle, brexpiprzole, brexpprazole, brxpiprazole, berxpiprazole, brepxiprazole, brexipprazole, brexpeprazole, Rexulti \\ \hline
Augmentation (antipsychotics) & cariprazine & cariprazin, cariprazene, cariprazne, cariprzine, carprazine, criprazine, acriprazine, cairprazine, careprazene, Vraylar \\ \hline
Augmentation (antipsychotics) & clozapine & clozapene, clozapinee, clozapin, clozapne, clozpine, clzapine, clazapine, cloazpine, clozaipne, Clozaril, Versacloz, FazaClo, cyclobenzaprine, clozapine \\ \hline
Augmentation (antipsychotics) & paliperidone & paliperdone, paliperidne, palipridone, palperidone, pliperidone, apliperidone, pailperidone, paleperedone, paliperidone \\ \hline
Augmentation (antipsychotics) & ziprasidone & ziprasdone, ziprasidne, ziprsidone, zprasidone, izprasidone, zeprasedone, ziparsidone, zipraisdone, Geodon, ziprasidone \\ \hline
Augmentation (antipsychotics) & iloperidone & iloperdone, iloperidne, ilopridone, ilperidone, eloperedone, ilaperidane, iloepridone, ilopeirdone \\ \hline
Augmentation (mood stabilizers) & lamotrigine & lamotrigene, lamatrigine, lamotriginee, lamotrgine, lamotrigin, lamotrigne, lamtrigine, lmotrigine, almotrigine, lamortigine, Lamictal, lamotrigine, lamotragine, lamotrogine \\ \hline
Augmentation (mood stabilizers) & pregabalin & pregablin, pregabaline, pregabaln, pregbalin, prgabalin, pergabalin, preagbalin, pregaablin, pregabailn, Lyrica, Lyrica CR, budweisers, pregabalin, lyrical \\ \hline
Augmentation (mood stabilizers) & lithium & lithim, lithum, lthium, ilthium, letheum, lihtium, lithimu, lithiom, Lithobid, Eskalith, Eskalith CR \\ \hline
Augmentation (mood stabilizers) & gabapentin & gabapentine, gabapentinne, gabapentn, gabapntin, gabpentin, gbapentin, agbapentin, gaabpentin, gabaepntin, gabapenitn, Neurontin, Gralise, Horizant, gabbies \\ \hline
Augmentation (mood stabilizers) & topiramate & topiramte, topirmate, topramate, tpiramate, otpiramate, tapiramate, toipramate, toperamate, topiramate, topamax \\ \hline
Augmentation (mood stabilizers) & valproate & valprate, valprote, vlproate, avlproate, valporate, valpraate, valpraote, valproaet, valproate, divalproex \\ \hline
Augmentation (mood stabilizers) & oxcarbazepine & oxcarbazepin, oxcarbazepne, oxcarbazpine, oxcarbzepine, oxcrbazepine, axcarbazepine, ocxarbazepine, oxacrbazepine, trileptal, oxcarbazepine \\ \hline
Augmentation (mood stabilizers) & carbamazepine & carbamazepin, carbamazepne, carbamazpine, carbamzepine, carbmazepine, crbamazepine, acrbamazepine, cabramazepine, carbamazepine \\ \hline
Augmentation (thyroid hormone) & liothyronine & liothyronine, triiodothyronine, T3, cytomel, liothyrone, lio-thyronine \\ \hline
Augmentation (thyroid hormone) & levothyroxine & levothyroxine, thyroxine, T4, L-thyroxine, synthroid, levoxyl, tirosint, unithroid, levothroid \\ \hline
MAOIs & tranylcypromine & tranylcipromine, tranylcypromin, tranylcyprmine, tranylcypromne, trnylcypromine, rtanylcypromine, tarnylcypromine, tranilcipromine, tranlycypromine, Parnate \\ \hline
MAOIs & phenelzine & phenelzin, phenalzine, phenelzne, phenlzine, phnelzine, hpenelzine, pehnelzine, pheenlzine, phenelizne, Nardil \\ \hline
MAOIs & selegiline & selegelin, selegilline, selegilin, selegilne, selegline, selgiline, slegiline, eslegiline, seelgiline, selegelene, Emsam \\ \hline
MAOIs & isocarboxazid & isocarboxasid, isocarboxyzid, iscarboxazid, isocarboxazd, isocarboxzid, isocarbxazid, isocrboxazid, esocarboxazed, ioscarboxazid, isacarbaxazid, Marplan \\ \hline
NaSSA & mirtazapine & mirtazepine, mirtazapene, mirtazapin, mirtazapne, mirtazpine, mirtzapine, mrtazapine, imrtazapine, mertazapene, miratzapine, Remeron, Avanza, Zispin, mirtazapine, mirtazipine,mirtanza \\ \hline
NDRI & bupropion & buproprion, bupropian, bpropion, bupropin, bupropon, buprpion, bopropion, bpuropion, buporpion, buprapian, Wellbutrin, Wellbutrin SR, Wellbutrin XL, Zyban, Aplenzin, bupropion, welbutrin \\ \hline
Neuroactive steroid / GABA-A modulator & zuranolone & zuranolone, zuranolone, zuranolon, SAGE-217, zurzuvae, sage217, Sage 217, SAGE1-217, zuranalone, zuranalone, zirzuvae, zurzurvae, gaba pill, zuralenone \\ \hline
Neuroactive steroid / GABA-A modulator & brexanolone & brexanolone, zulresso, allopregnanolone, sage-547, Brexalone, Brexanoline, breaxnolone, brexanalone, allpregnanolone, allpreg, allopreg, allopregenanalone \\ \hline
Neuromodulation & deep transcranial magnetic stimulation & deep transcranial magnetic stimulation, deep TMS, dTMS, H-coil, H coil, BrainsWay, dtms, d-tms, H1 coil, H-1 coil, H7 coil, H4, H-shaped coil, helmet \\ \hline
Neuromodulation & electroconvulsive therapy & electroconvulsive therapy, ECT, electroshock, shock therapy, mECT, maintenance ECT, electro convulsive, bitemporal, bifrontal, right unilateral, RUL, bilateral, unilateral, ultrabrief pulse, brief pulse, eletroconvulsive, electroconvulsive theraphy, electroconvulsive treatment \\ \hline
Neuromodulation & intermittent theta burst stimulation & intermittent theta burst stimulation, iTBS, theta burst, theta burst stimulation, TBS, tbs, cTBS, continuous theta burst, Express TMS, 3-minute protocol, 9-minute protocol, MagVenture \\ \hline
Neuromodulation & repetitive transcranial magnetic stimulation & transcranial magnetic stimulation, TMS, rTMS, repetitive TMS, repetitive transcranial magnetic stimulation, TMS therapy, rdTMS, NeuroStar, Neuronetics, Magstim Horizon, Magstim Horizon 3.0, StimGuide, figure-8 coil, CloudTMS, 19-minute protocol, 20-minute protocol, 37-minute protocol \\ \hline
NMDA / Rapid-acting & ketamine & ketamin, ketamime, ketamne, ketmine, ktamine, ektamine, keatmine, ketaimne, ketamene, Ketalar, special k, vitamin k, k, ketamine \\ \hline
NMDA / Rapid-acting & esketamine & esketamin, s-ketamine, esketamne, esketmine, esktamine, eksetamine, esektamine, eskeatmine, esketaimne, Spravato, s ketamine \\ \hline
NMDA / Rapid-acting & dextromethorphan-bupropion & dextrmethorphan-bupropion, dextromethorphan-bpropion, dextromethorphan-bupropin, dextromethorphan-bupropon, dextromethorphan-buprpion, dextromethorphn-bupropion, dextromethrphan-bupropion, dextromthorphan-bupropion, Auvelity, Auvelityy, Auvility, Avelity, Auvelty, Auvality, Alvelity, AXS-05, dextromethorphan bupropion, dextromethorphan + bupropion \\ \hline
Other / Misc & psilocybin & psilcybin, psilocybn, pslocybin, pislocybin, pselocyben, psilacybin, psilcoybin, psilocbyin, Psilocybin, shrooms, magic mushrooms, mushrooms, mushroom, shroom \\ \hline
Other / Misc & thyroid & thyrid, thyrod, htyroid, thiroid, thryoid, thyorid, thyraid, thyriod, thyroxine \\ \hline
Other / Misc & sam-e & sm-e, asm-e, sa-me, sam-i, same-, sma-e, som-e, sam--e \\ \hline
Other / Misc & omega-3 & omeg-3, omga-3, amega-3, moega-3, oemga-3, omeag-3, omeg-a3, omega3- \\ \hline
SNRIs & venlafaxine & venlafexine, venafaxine, venlafaxin, venlafaxne, venlafxine, venlfaxine, vnlafaxine, evnlafaxine, velnafaxine, venalfaxine, Effexor, Effexor XR, venlafaxine \\ \hline
SNRIs & duloxetine & duloxitine, duloxatine, duloxotine, dloxetine, duloxetin, duloxetne, duloxtine, dulxetine, dluoxetine, doloxetine, dulaxetine, Cymbalta, Irenka, duloxetine \\ \hline
SNRIs & desvenlafaxine & desvenlafexine, desvenlafaxin, desvenlafaxne, desvenlafxine, desvenlfaxine, desvnlafaxine, dsvenlafaxine, desevnlafaxine, desvelnafaxine, Pristiq, desvenlafaxine \\ \hline
SNRIs & levomilnacipran & levmilnacipran, levomilnaciprn, levomilnacpran, levomilncipran, levomlnacipran, lvomilnacipran, elvomilnacipran, leovmilnacipran \\ \hline
SNRIs & milnacipran & milnaciprn, milnacpran, milncipran, mlnacipran, imlnacipran, melnacepran, milancipran, milnaciparn \\ \hline
SSRIs & fluoxetine & fluoxitine, fluxetine, flooxetine, floxetine, fluoxetin, fluoxetne, fluoxtine, flouxetine, fluaxetine, Prozac, Sarafem, fluoxetine, prozak \\ \hline
SSRIs & sertraline & sertaline, sertralin, sertralene, sertralne, sertrline, srtraline, esrtraline, serrtaline, sertarline, sertrailne, Zoloft, sertraline, setraline \\ \hline
SSRIs & escitalopram & escitolopram, es-citalopram, escitaloprm, escitalpram, escitlopram, esctalopram, ecsitalopram, escetalopram, esciatlopram, escitalapram, Lexapro, escitalopram \\ \hline
SSRIs & paroxetine & paroxitine, paxotine, paroxetin, paroxetne, paroxtine, parxetine, proxetine, aproxetine, paorxetine, paraxetine, Paxil, Paxil CR, Brisdelle, paroxetine, promethazine \\ \hline
SSRIs & citalopram & citolopram, citralopram, citaloprm, citalpram, citlopram, ctalopram, cetalopram, ciatlopram, citalapram, citaloparm, Celexa, citalopram \\ \hline
SSRIs & fluvoxamine & fluvoxamin, fluvoxamne, fluvoxmine, fluvxamine, flvoxamine, flovoxamine, fluovxamine, fluvaxamine \\ \hline
Stimulants / ADHD adjuncts & amphetamine-dextroamphetamine & amphetamin-dextroamphetamine, amphetamine-dextramphetamine, amphetamine-dextroamphetamin, amphetamine-dextroamphetamne, amphetamine-dextroamphetmine, amphetamine-dextroamphtamine, amphetamine-dextromphetamine, amphetamine-dxtroamphetamine, adderall, amphetamine, adderal, amphetamines, aderall, adderrall \\ \hline
Stimulants / ADHD adjuncts & methylphenidate & methylphenadate, methylphenidete, methylphendate, methylphenidte, methylphnidate, mthylphenidate, emthylphenidate, mehtylphenidate, methilphenidate, methlyphenidate, Ritalin, Concerta, Metadate, Daytrana, Quillivant XR, Aptensio XR, Jornay PM, rits, methylphenidate \\ \hline
Stimulants / ADHD adjuncts & lisdexamfetamine & lisdexamphetamine, lisdexanfetamine, lisdexamfetamin, lisdexamfetamne, lisdexamfetmine, lisdexamftamine, lisdexmfetamine, lisdxamfetamine, lsdexamfetamine, ilsdexamfetamine, Vyvanse, vyvance, lisdexamfetamine \\ \hline
Stimulants / ADHD adjuncts & modafinil & modafanil, modafinill, mdafinil, modafinl, modafnil, modfinil, madafinil, mdoafinil, moadfinil, modafenel, Provigil, modafinil \\ \hline
Stimulants / ADHD adjuncts & atomoxetine & atomoxitine, atomoxatine, atmoxetine, atomoxetin, atomoxetne, atomoxtine, atomxetine, aotmoxetine, atamaxetine, atmooxetine, Strattera, atomoxetine, straterra \\ \hline
Stimulants / ADHD adjuncts & dextroamphetamine & dextramphetamine, dextroamphetamin, dextroamphetamne, dextroamphetmine, dextroamphtamine, dextromphetamine, dxtroamphetamine, detxroamphetamine, Dexedrine, Zenzedi, ProCentra, dex, dexies, dextroamphetamine, dexamphetamine \\ \hline
Stimulants / ADHD adjuncts & clonidine & clnidine, clondine, clonidin, clonidne, clanidine, clnoidine, cloindine, clondiine, Kapvay, Catapres, clonidine, clonodine \\ \hline
Stimulants / ADHD adjuncts & armodafinil & armodafanil, armodafinill, armdafinil, armodafinl, armodafnil, armodfinil, amrodafinil, armadafinil, armdoafinil, armoadfinil, Nuvigil, armodafinil \\ \hline
Stimulants / ADHD adjuncts & guanfacine & ganfacine, guanfacin, guanfacne, guanfcine, gunfacine, gaunfacine, goanfacine, guafnacine, Intuniv, Tenex \\ \hline
TCAs & clomipramine & clomipramene, clomipraminee, clmipramine, clomipramin, clomipramne, clomiprmine, clompramine, clamipramine, clmoipramine, cloimpramine, Anafranil, clomipramine \\ \hline
TCAs & amitriptyline & amitriptline, amitriptilline, amitriptylinee, amitriptylin, amitriptylne, amitrptyline, amtriptyline, aimtriptyline, ametreptylene, amirtiptyline, amitirptyline, Elavil, Endep, amitriptyline, amitryptiline, amitryptaline \\ \hline
TCAs & nortriptyline & nortriptline, nortriptiline, nortriptylinee, nortriptylin, nortriptylne, nortrptyline, nrtriptyline, nartriptyline, norrtiptyline, nortirptyline, nortreptylene, Pamelor, Aventyl, nortriptyline \\ \hline
TCAs & imipramine & imipramene, imipraminee, imipramin, imipramne, imiprmine, impramine, emepramene, iimpramine, imiparmine, imipraimne, Tofranil, imipramine \\ \hline
TCAs & doxepin & doxepine, doxapin, doxepn, doxpin, dxepin, daxepin, doexpin, doxeipn, doxepen, doxepni, Silenor, Adapin, Sinequan, doxepin, dosulepin \\ \hline
TCAs & trimipramine & trimipramene, trimipraminee, trimipramin, trimipramne, trimiprmine, trimpramine, trmipramine, rtimipramine, tirmipramine, tremepramene, Surmontil \\ \hline
TCAs & protriptyline & protriptline, protriptiline, protriptylin, protriptylne, protrptyline, prtriptyline, portriptyline, pratriptyline, prortiptyline, protirptyline, Vivactil \\ \hline
TCAs & amoxapine & amoxapin, amoxapne, amoxpine, amxapine, amaxapine, amoaxpine, amoxaipne, amoxapene \\ \hline
TCAs & desipramine & desipramin, desipramne, desiprmine, despramine, dsipramine, deispramine, desepramene, desiparmine \\ \hline

\end{longtable}

%\end{landscape}
%\end{landscape}

% Table S2. Subreddits

\begin{longtable}{p{0.24\linewidth}p{0.28\linewidth}p{0.42\linewidth}}
\caption{\textbf{Subreddits queried for TRD-related data collection.}
Subreddits are grouped by the categories used during data collection: general mental health/advice forums, medication-focused communities, and comorbid-condition communities. Notes indicate communities flagged as NSFW or professional-perspective forums.}
\label{tab:subreddits}\\
\toprule
Category & Subreddit & Notes \\
\midrule
\endfirsthead

\toprule
Category & Subreddit & Notes \\
\midrule
\endhead

\midrule
\multicolumn{3}{r}{\footnotesize Continued on next page.}\\
\endfoot

\bottomrule
\endlastfoot

General posts & \texttt{r/mentalhealth} &  \\
General posts & \texttt{r/MentalHealthUK} &  \\
General posts & \texttt{r/depressionregimens} &  \\
General posts & \texttt{r/depression} &  \\
General posts & \texttt{r/mentalillness} &  \\
General posts & \texttt{r/depression\_help} &  \\
General posts & \texttt{r/AskPsychiatry} &  \\
General posts & \texttt{r/askatherapist} &  \\
General posts & \texttt{r/AskDocs} &  \\
General posts & \texttt{r/StackAdvice} & NSFW; self-reports of TRD conditions and prescriptions noted. \\
General posts & \texttt{r/Drugs} & NSFW; self-reports of TRD conditions and prescriptions noted. \\
General posts & \texttt{r/therapists} & Therapists discuss clients with TRD. \\
General posts & \texttt{r/Psychiatry} & Perspectives of psychiatrists. \\
General posts & \texttt{r/therapy} &  \\

Medication-focused & \texttt{r/antidepressants} &  \\
Medication-focused & \texttt{r/Spravato} &  \\
Medication-focused & \texttt{r/seroquelmedication} &  \\
Medication-focused & \texttt{r/MAOIs} &  \\
Medication-focused & \texttt{r/KetamineTherapy} &  \\
Medication-focused & \texttt{r/TherapeuticKetamine} &  \\
Medication-focused & \texttt{r/lamictal} &  \\
Medication-focused & \texttt{r/trintellix} &  \\
Medication-focused & \texttt{r/gabapentin} &  \\
Medication-focused & \texttt{r/prozac} &  \\
Medication-focused & \texttt{r/cymbalta} &  \\

Comorbid conditions & \texttt{r/BipolarReddit} &  \\
Comorbid conditions & \texttt{r/bipolar} &  \\
Comorbid conditions & \texttt{r/AnxietyDepression} &  \\
Comorbid conditions & \texttt{r/ADHD} &  \\

\end{longtable}

% Table S3. TRD-identification keywords/phrases

\begin{longtable}{p{0.26\linewidth}p{0.68\linewidth}}
\caption{\textbf{Keyword/phrase lexicon used to identify TRD-related posts.}
Phrases were used as heuristic indicators of TRD contexts, including formal clinical terms, abbreviations, and common colloquial expressions describing repeated treatment failure.}
\label{tab:keywords}\\
\toprule
Category & Keyword/phrase \\
\midrule
\endfirsthead

\toprule
Category & Keyword/phrase \\
\midrule
\endhead

\midrule
\multicolumn{2}{r}{\footnotesize Continued on next page.}\\
\endfoot

\bottomrule
\endlastfoot

Core clinical terms & \texttt{treatment-resistant depression} \\
Core clinical terms & \texttt{treatment resistant depression} \\
Core clinical terms & \texttt{treatment-resistant MDD} \\
Core clinical terms & \texttt{difficult-to-treat depression} \\
Core clinical terms & \texttt{difficult-to-treat MDD} \\

Abbreviations \& related terms & \texttt{TRD} \\
Abbreviations \& related terms & \texttt{DTD} \\
Abbreviations \& related terms & \texttt{refractory depression} \\
Abbreviations \& related terms & \texttt{SSRI-resistant depression} \\
Abbreviations \& related terms & \texttt{incurable depression} \\

Colloquialisms & \texttt{failed meds} \\
Colloquialisms & \texttt{failed two meds} \\
Colloquialisms & \texttt{medication roulette} \\
Colloquialisms & \texttt{nothing's working} \\
Colloquialisms & \texttt{maximum dose} \\
Colloquialisms & \texttt{tried every} \\
Colloquialisms & \texttt{tried everything} \\

\end{longtable}

\begin{longtable}{p{0.28\textwidth}rrrccc}
\caption{\textbf{Therapies with significant asymmetry between positive and negative evaluations among non-neutral mentions (two-sided exact binomial test).} The positive share is the proportion of positive mentions among non-neutral mentions; 95\% exact binomial confidence intervals are shown. Benjamini--Hochberg adjusted $p_{\mathrm{FDR}}$ values are reported.}
\label{tab:full_medication_pfdr}\\
\hline
Medication & Positive $n$ & Negative $n$ & Non-neutral $n$ & $\hat{p}$ (Positive) & 95\% CI for $\hat{p}$ & $p_{\mathrm{FDR}}$ \\
\hline
\endfirsthead
\hline
Medication & Positive $n$ & Negative $n$ & Non-neutral $n$ & $\hat{p}$ (Positive) & 95\% CI for $\hat{p}$ & $p_{\mathrm{FDR}}$ \\
\hline
\endhead
\hline
\multicolumn{7}{r}{\footnotesize Continued on next page.}\\
\endfoot
\hline
\endlastfoot
Lisdexamfetamine & 179 & 44 & 223 & 0.803 & [0.744, 0.853] & $1.25\times 10^{-18}$ \\
Sertraline & 52 & 174 & 226 & 0.230 & [0.177, 0.291] & $5.48\times 10^{-15}$ \\
Venlafaxine & 46 & 158 & 204 & 0.225 & [0.170, 0.289] & $3.59\times 10^{-14}$ \\
Mirtazapine & 49 & 147 & 196 & 0.250 & [0.191, 0.317] & $2.72\times 10^{-11}$ \\
Pregabalin & 97 & 28 & 125 & 0.776 & [0.693, 0.846] & $6.38\times 10^{-9}$ \\
Citalopram & 11 & 47 & 58 & 0.190 & [0.099, 0.314] & $2.57\times 10^{-5}$ \\
Repetitive Transcranial Magnetic Stimulation & 97 & 175 & 272 & 0.357 & [0.300, 0.417] & $2.83\times 10^{-5}$ \\
Ketamine & 496 & 370 & 866 & 0.573 & [0.539, 0.606] & $1.99\times 10^{-4}$ \\
Fluoxetine & 55 & 109 & 164 & 0.335 & [0.264, 0.413] & $2.53\times 10^{-4}$ \\
Bupropion & 127 & 201 & 328 & 0.387 & [0.334, 0.442] & $3.95\times 10^{-4}$ \\
Electroconvulsive Therapy & 132 & 204 & 336 & 0.393 & [0.340, 0.447] & $7.01\times 10^{-4}$ \\
Vilazodone & 7 & 30 & 37 & 0.189 & [0.080, 0.352] & $1.21\times 10^{-3}$ \\
Esketamine & 197 & 131 & 328 & 0.601 & [0.545, 0.654] & $1.86\times 10^{-3}$ \\
Amphetamine--dextroamphetamine & 169 & 110 & 279 & 0.606 & [0.546, 0.663] & $2.68\times 10^{-3}$ \\
Aripiprazole & 79 & 127 & 206 & 0.383 & [0.317, 0.454] & $5.12\times 10^{-3}$ \\
Nefazodone & 10 & 0 & 10 & 1.000 & [0.692, 1.000] & $9.28\times 10^{-3}$ \\
Phenelzine & 100 & 62 & 162 & 0.617 & [0.538, 0.692] & $1.57\times 10^{-2}$ \\
Fluvoxamine & 4 & 17 & 21 & 0.190 & [0.054, 0.419] & $2.88\times 10^{-2}$ \\
Quetiapine & 43 & 73 & 116 & 0.371 & [0.283, 0.465] & $2.88\times 10^{-2}$ \\
Paroxetine & 13 & 31 & 44 & 0.295 & [0.168, 0.452] & $3.63\times 10^{-2}$ \\
\end{longtable}

\begin{figure*}[ht]
\includegraphics[scale=0.5]{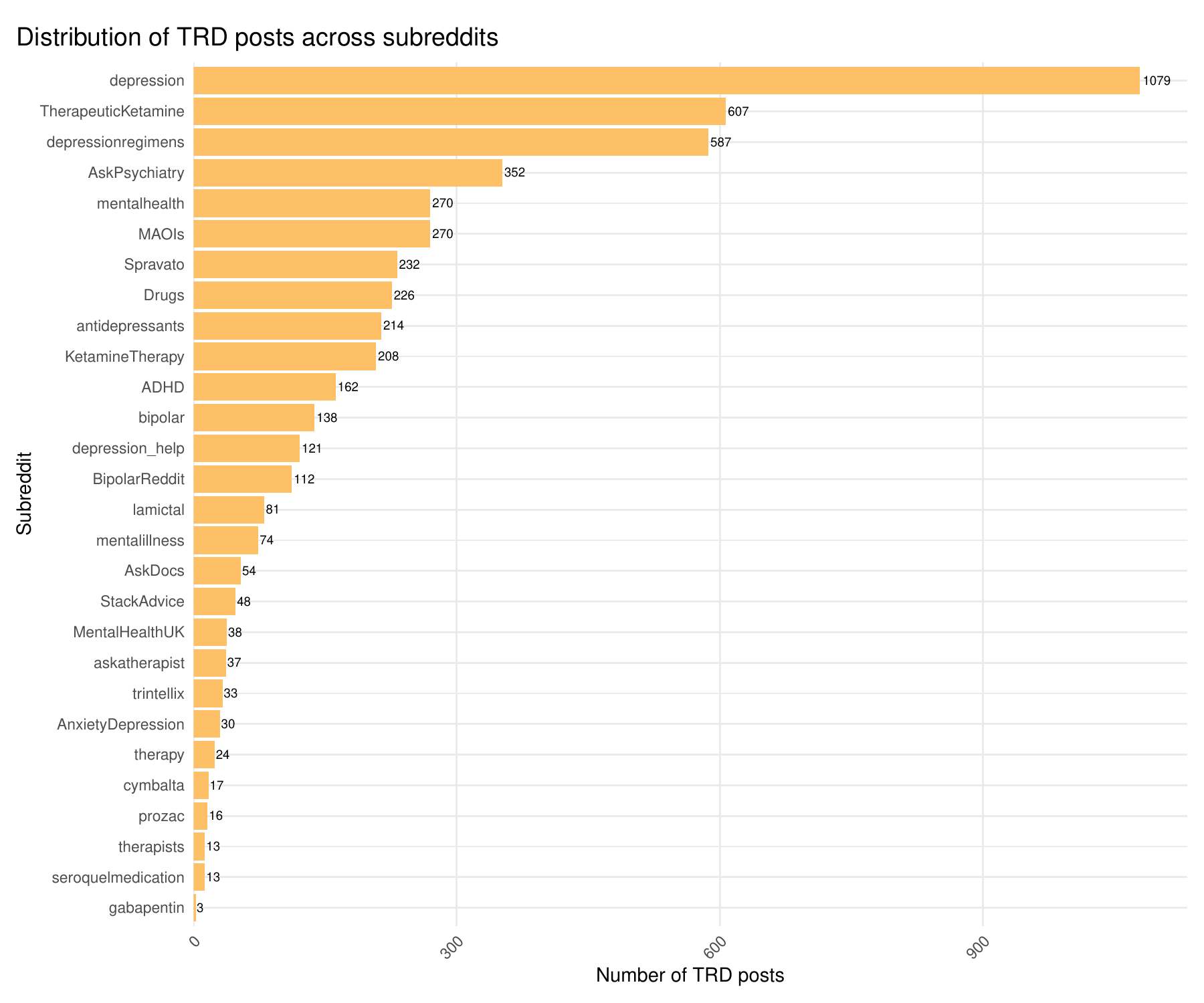}
\caption{Distribution of TRD-relevant posts across subreddits. Horizontal bars report the number of posts in each of the 28 included subreddits; subreddits are ordered by post count (highest to lowest). Values at bar ends give exact counts, highlighting concentration in general depression forums (e.g., r/depression: 1,079 posts) and therapy-focused communities (e.g., r/TherapeuticKetamine: 607; r/MAIOs: 587).}
\label{fig:subreddits}
\end{figure*}

\begin{figure*}[ht]
\includegraphics[scale=0.5]{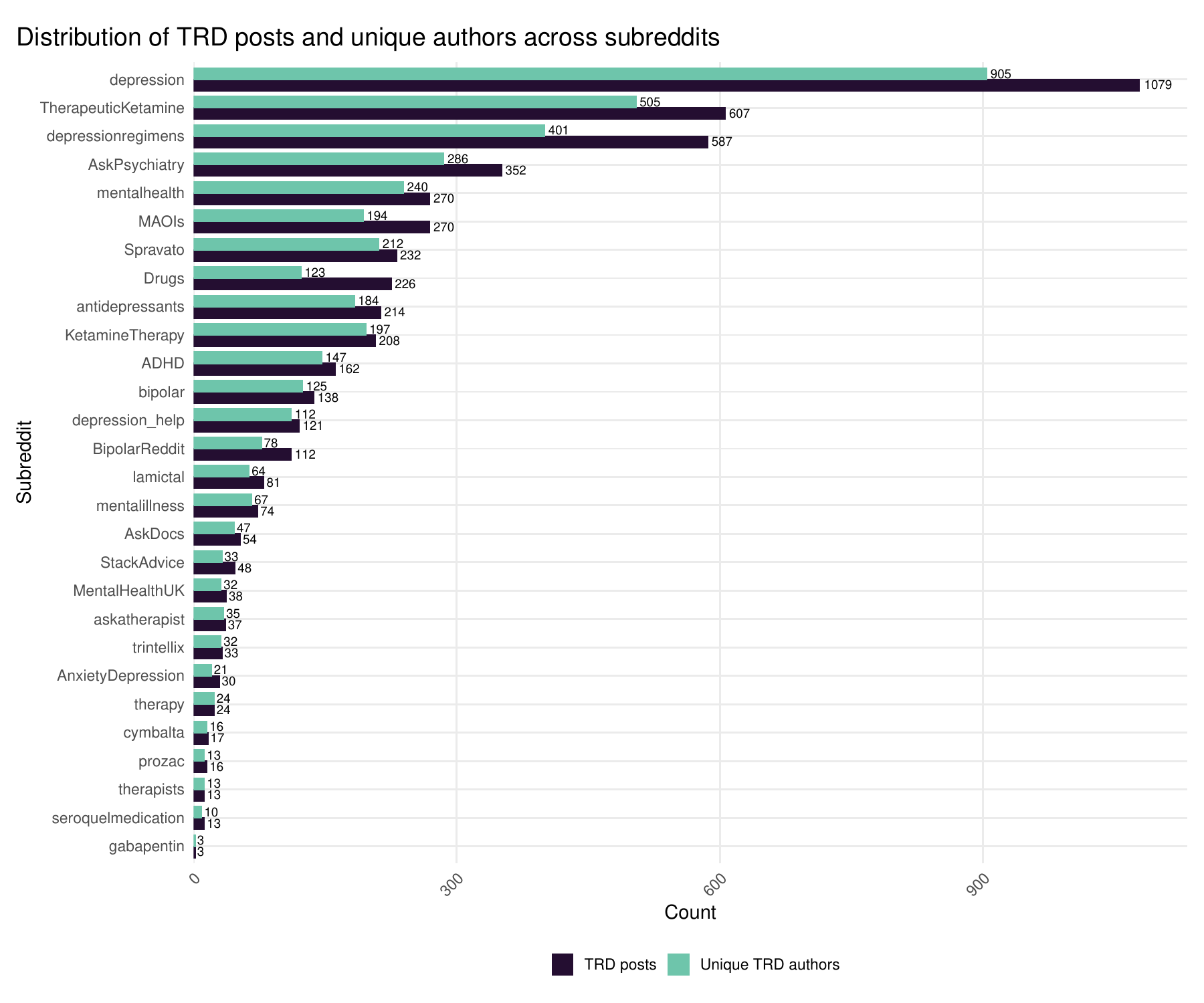}
\caption{Subreddit-level volume and contributor breadth of TRD-relevant discussion. For each subreddit, paired bars show the number of TRD-relevant posts (dark) and the number of unique authors contributing those posts (teal), with subreddits ordered by post count. Counts at bar ends provide exact values (e.g., r/depression: 1,079 posts from 905 unique authors), enabling comparison of discussion volume versus concentration among repeat contributors.}
\label{fig:subreddits_author}
\end{figure*}

\begin{figure*}[ht]
\includegraphics[scale=0.5]{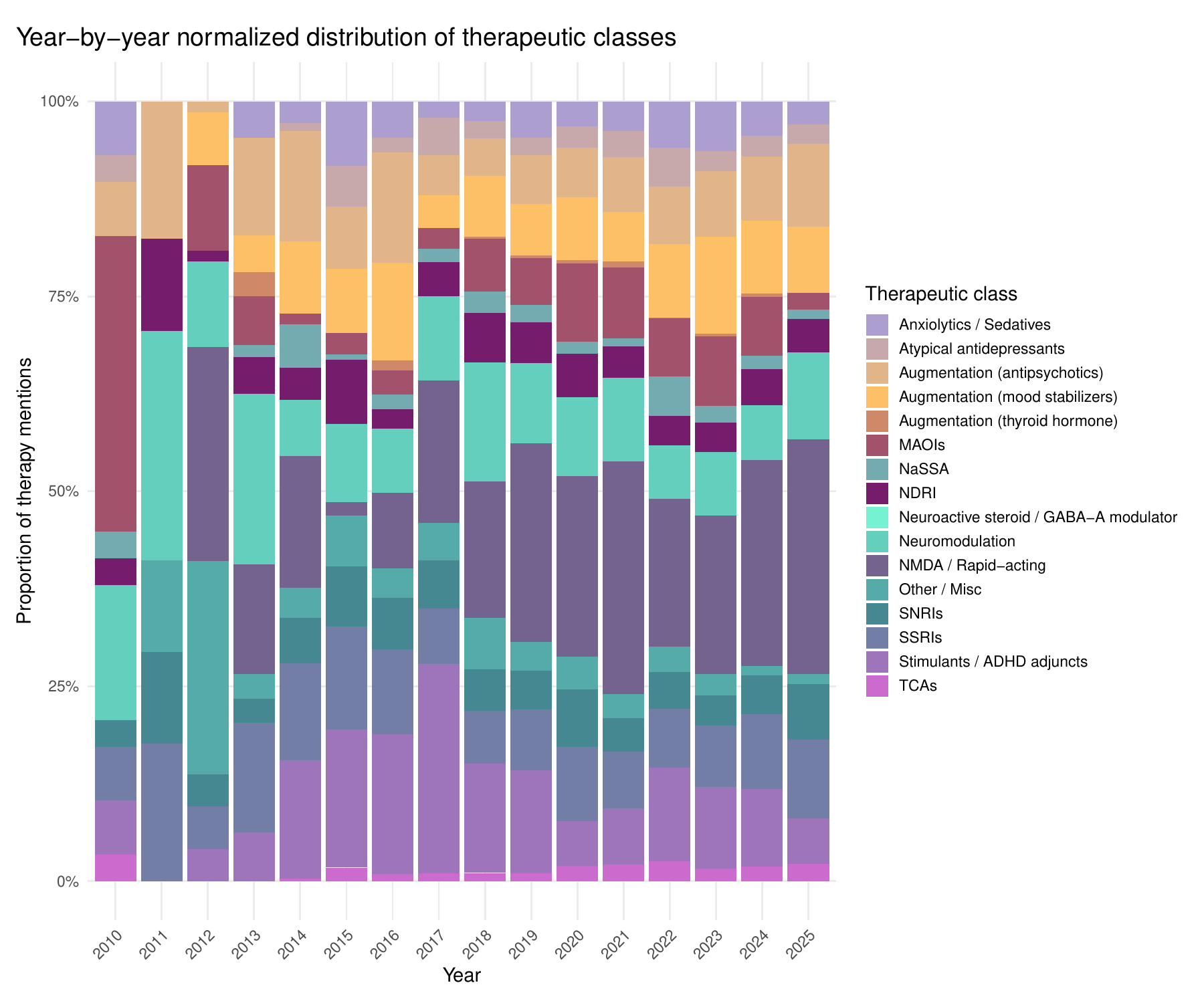}
\caption{Year-by-year composition of medication mentions by therapeutic class in TRD-relevant posts (2010–2025). Each stacked bar corresponds to a calendar year and shows the proportion of medication mentions assigned to each therapeutic class). For each year, the denominator is the total number of medication mentions in that year, and each segment numerator is the number of mentions in that therapeutic class in that same year. Within each year, segments sum to 100\%, hence the figure reflects changes in the relative mix of classes over time.}
\label{fig:class_yearly}
\end{figure*}

\renewcommand{\arraystretch}{1.12}
\setlength{\LTpre}{0pt}
\setlength{\LTpost}{0pt}

\begin{longtable}{p{0.26\linewidth}p{0.68\linewidth}}
\caption{\textbf{Keyword/phrase lexicon used to identify TRD-related posts.}
Phrases were used as heuristic indicators of TRD contexts, including formal clinical terms, abbreviations, and common colloquial expressions describing repeated treatment failure.}
\label{tab:keywords}\\
\toprule
Category & Keyword/phrase \\
\midrule
\endfirsthead

\toprule
Category & Keyword/phrase \\
\midrule
\endhead

\midrule
\multicolumn{2}{r}{\footnotesize Continued on next page.}\\
\endfoot

\bottomrule
\endlastfoot

Core clinical terms & \texttt{treatment-resistant depression} \\
Core clinical terms & \texttt{treatment resistant depression} \\
Core clinical terms & \texttt{treatment-resistant MDD} \\

Abbreviations \& related terms & \texttt{TRD} \\
Abbreviations \& related terms & \texttt{refractory depression} \\
Abbreviations \& related terms & \texttt{SSRI-resistant depression} \\
Abbreviations \& related terms & \texttt{incurable depression} \\

Colloquialisms & \texttt{failed meds} \\
Colloquialisms & \texttt{failed two meds} \\
Colloquialisms & \texttt{medication roulette} \\
Colloquialisms & \texttt{nothing's working} \\
Colloquialisms & \texttt{maximum dose} \\
Colloquialisms & \texttt{tried everything} \\

\end{longtable}

\begingroup
\small
\renewcommand{\arraystretch}{1.15}
\begin{longtable}{>{\raggedright\arraybackslash}p{0.17\textwidth}
                  >{\raggedright\arraybackslash}p{0.60\textwidth}
                  >{\raggedright\arraybackslash}p{0.11\textwidth}
                  >{\raggedleft\arraybackslash}p{0.08\textwidth}}
\caption{\textbf{Paraphrased first-person excerpts illustrating mention-level sentiment classifications for each therapy.} For each generic-name therapy, we show up to two examples selected as the highest-confidence predictions from the DeBERTa-based classifier (typically one \textsc{Positive} and one \textsc{Negative}; when only one polarity was available, a \textsc{Neutral} example is shown instead). Excerpts are researcher-generated paraphrases (not verbatim Reddit text) that preserve the original first-person perspective while removing identifying details. Sentiment labels reflect the evaluation of the \emph{target therapy} in its local context window (the sentence containing the mention with surrounding context; rather than the overall post tone). $P(\hat{y})$ denotes the model's predicted probability for the assigned label.}\label{tab:sentiment_examples_by_therapy}\\
\toprule
Therapy (generic name) & Paraphrased first-person example (therapy in \textbf{bold}) & Label & $P(\hat{y})$ \\
\midrule
\endfirsthead

\toprule
Therapy (generic name) & Paraphrased first-person example (therapy in \textbf{bold}) & Label & $P(\hat{y})$ \\
\midrule
\endhead

\midrule
\multicolumn{4}{r}{\emph{Continued on next page.}}\\
\endfoot

\bottomrule
\endlastfoot

alprazolam & I have tried 16 different antidepressants and failed. I was prescribed \textbf{alprazolam} at age 16 which caused me to develop a dependence (successfully came off that except PRN last year). Last year, my doctor prescribed TMS and Ketamine therapy. & \textsc{Negative} & 0.989 \\
alprazolam & *drinking alcohol? no problemo!, you can spend as much money as you want on that, as long as big gov is making some money, pick up some smokes while your at it!, taking some diazepam/\textbf{alprazolam}?* where is your prescription, you are a drug addict, why are you self medicating, you need a fine, criminal record ... & \textsc{Neutral} & 0.989 \\
alprazolam & Not exactly where I want to be but an improvement for sure. My Dr prescribed .5 \textbf{alprazolam} for anxiety as needed. I have found it sooo helpful. & \textsc{Positive} & 0.999 \\
amitriptyline & Has anyone had a similar experience? or had any luck with \textbf{amitriptyline}? I just hate feeling like this. & \textsc{Negative} & 0.995 \\
amitriptyline & Have you tried the tcas? specifically, \textbf{amitriptyline}, imipramine and clomipramine? 3. & \textsc{Neutral} & 0.990 \\
amitriptyline & I’ve wondered why tricyclic medications have never been brought up as a possible choice for me. I do take 25mg of \textbf{amitriptyline} at bedtime for migraine prevention; it has been extremely helpful for this. I have never noticed any adverse effects but I know it’s a low dose. & \textsc{Positive} & 0.999 \\
amoxapine & \textbf{amoxapine} \# Norepinephrine reuptake inhibitor (NRIs) **Extract from “Stahl's Essential Psychopharmacology recommendations”**: \&gt;**Use with caution with MAOIs, including 14 days after MAOIs are stopped (for the expert)**. \&amp;\#x200B; [**MAOIs: Swapping and Combining**]( (article) claims that atomoxetine can be safely combined with MAOIs: \&gt;**Other safe drugs** (with insignificant SRI potency): […] **atomoxetine** \&amp;\#x200B; Nevertheless I think that ... & \textsc{Neutral} & 0.981 \\
amoxapine & \textbf{amoxapine} \# Norepinephrine reuptake inhibitor (NRIs) **Extract from “Stahl's Essential Psychopharmacology recommendations”**: \&gt;**Use with caution with MAOIs, including 14 days after MAOIs are stopped (for the expert)**. \&amp;\#x200B; [**MAOIs: Swapping and Combining**]( (article) claims that atomoxetine can be safely combined with MAOIs: \&gt;**Other safe drugs** (with insignificant SRI potency): […] **atomoxetine** \&amp;\#x200B; Nevertheless I think that ... & \textsc{Neutral} & 0.981 \\
amoxapine & \&amp;nbsp; 54m, adhd-I, depression, anxiety *many* medical issues because of treatment-resistant depression and because I quickly developed tolerance to adderall, I was referred to a psychopharmacologist. I asked whether I could do something for amphetamine similar to how naltrexone works with opioids, and he liked the idea and put me on \textbf{amoxapine} to kill three ... & \textsc{Positive} & 0.702 \\
amphetamine dextroamphetamine & I can barely articulate my thoughts half the time. TL;DR: \textbf{amphetamine dextroamphetamine} has affected my memory, I feel dumber on it, and have issues concentrating and was prescribed it for TRD. this is a net negative in my life and I am experiencing anhedonia. & \textsc{Negative} & 0.998 \\
amphetamine dextroamphetamine & Xanax stimulants 1. various \textbf{amphetamine dextroamphetamine} salts 2. various methlyphendiate formulations 3. & \textsc{Neutral} & 0.990 \\
amphetamine dextroamphetamine & \textbf{amphetamine dextroamphetamine} XR has been a game changer! I am diagnosed with treatment resistant depression and anxiety. & \textsc{Positive} & 1.000 \\
aripiprazole & Med change? I am on depakote, cymbalta, and \textbf{aripiprazole}, and they're not working. I am having incredible mood swings where I go from perky and happy to almost in tears in minutes, and my concentration is going to hell. & \textsc{Negative} & 0.998 \\
aripiprazole & More or less clear by now that I have trd with some mood features. I am on 40mg prozac, 2mg \textbf{aripiprazole} and going on lamotragine soon. thinking about throwing inositol into the mix. & \textsc{Neutral} & 0.990 \\
aripiprazole & Does it help? I’m taking 60mg of Lexapro, 400mg of Bupropion SR, and 10mg of \textbf{aripiprazole}. This is the perfect combo for me. & \textsc{Positive} & 0.999 \\
armodafinil & I was previously on adderall ir (20mg three times a day) for add as well as chronic fatigue; but was told to discontinue due to contraindication. I was then prescribed \textbf{armodafinil} - not the first time I have taken this medication. it did not work then, and it certainly isn't working now. & \textsc{Negative} & 0.930 \\
armodafinil & I really want to ask my pdoc about modafinil— I was thinking two 100mg tablets qam and one 100mg tablet qnoon after that. I have been rx'd \textbf{armodafinil} [nuvigil] before in the past at 250mg. I remember some type of benefit, but that is really all I remember— I think 60mg amphetamine salts ir (40mg ... & \textsc{Neutral} & 0.986 \\
armodafinil & I have always been a good test taker and this test was never going to accurately model the type of distractors that hijack my attention in real life. When I complained of fatigue and difficulty concentrating on work to my new pdoc, he prescribed \textbf{armodafinil}. I serendipitously discovered that it completely eliminated all of my excessive ... & \textsc{Positive} & 0.994 \\
asenapine & Pdoc says treatment resistant depression... ssri/snri = because mania seroquel, latuda, \textbf{asenapine} = all caused nueorleptic malignant syndrome so atypicals are out. lamictal, tripletal, valproate = mixed episode and rage. & \textsc{Neutral} & 0.989 \\
asenapine & Pdoc says treatment resistant depression... ssri/snri = because mania seroquel, latuda, \textbf{asenapine} = all caused nueorleptic malignant syndrome so atypicals are out. lamictal, tripletal, valproate = mixed episode and rage. & \textsc{Neutral} & 0.989 \\
asenapine & \textbf{asenapine} Just want to get your opinions on this and any comments if you’ve tried the same medications. The medications I’ve tried that have worked out (slightly or completely) have been: • Medical Marijuana: Can’t use because I’m in sobriety, it worked out for anxiety, social anxiety, and insomnia • Gabapentin: The second time I got a tiny ... & \textsc{Positive} & 0.908 \\
atomoxetine & I have just spent over a month on a new medication regime, only to be disappointed yet again. I do not feel like abilify is doing anything for me, and the \textbf{atomoxetine} given for my adhd hasn't helped, either. in fact, the past few weeks have been utter hell. & \textsc{Negative} & 0.997 \\
atomoxetine & Prescribed \textbf{atomoxetine} without an adhd diagnosis? hello. & \textsc{Neutral} & 0.989 \\
atomoxetine & \textbf{atomoxetine}, reboxetine or bupropion for difficulty initiating tasks and anergia? between atomoxetine, reboxetine or bupropion, which one is the most effective at treating difficulty initiating tasks, anergia and mild cognitive impairment? & \textsc{Positive} & 0.888 \\
brexpiprazole & Most things either made me worse or numbed me out until I was a zombie. The only exception was \textbf{brexpiprazole}, which gave me one day of relief—a glimpse of motivation, reduced rumination, and a sense of peace I hadn’t felt in years. But after that, it turned into a complete nightmare: severe insomnia, agitation, rage, ... & \textsc{Negative} & 0.992 \\
brexpiprazole & I’ve tried ability. \textbf{brexpiprazole}. wellbuitrin etc. & \textsc{Neutral} & 0.988 \\
brexpiprazole & As of about 10 months ago, I am in remission from lifelong, severe, treatment-resistant depression thanks to tms therapy. I take a couple of medications (low doses of trintellix and \textbf{brexpiprazole}) as well that manage any lingering symptoms pretty well, but my inability to focus, forgetfulness, etc. seems to be in sharper relief now that the ... & \textsc{Positive} & 0.998 \\
bupropion & Prozac and \textbf{bupropion} not working what’s next? I have been on zoloft and luvox in the past but none of them really worked out for my depression. & \textsc{Negative} & 0.999 \\
bupropion & Any other ideas? should I ask about modafinil and, if so, taper the \textbf{bupropion} qam? for anxiety, perhaps ask for klonopin 2mg over the course of a day along with the gabapentin rx? & \textsc{Neutral} & 0.989 \\
bupropion & My med combo is working great Vortioxetine + \textbf{bupropion} + Vyvanse. !! & \textsc{Positive} & 0.999 \\
buspirone & Maybe not many of the newer ones. \textbf{buspirone} doesn’t do anything for the anxiety. What has been the magical combo for you??? & \textsc{Negative} & 0.998 \\
buspirone & Wellbutrin 2. \textbf{buspirone} 3. trazadone clinical treatment 1. & \textsc{Neutral} & 0.989 \\
buspirone & All symptoms persisted. I eventually discovered the critical role of the 5-HT1A receptor (regulating mood, anxiety, memory, and most importantly in my case: **REM sleep.** \textbf{buspirone}, a 5-HT1A partial agonist, taken before bed, has significantly improved my ability to stage into REM sleep. The change was profound: my brain fog lifted entirely, my mood became ... & \textsc{Positive} & 0.999 \\
carbamazepine & Treatment resistant depression/anxiety I have tried a whole bunch of medications but I am still suicidal no matter what I have tried. this includes zoloft lithium, abilify, lamictal, \textbf{carbamazepine}, depakote, seroquel, olanzapine, prozac+olanzapine, gabapentin, riseperidal, wellbutrin, ritalin, guanfacine, cymbalta, and trintellix. I find that my baseline when I am off of medications is severe waves of debilitating anxiety ... & \textsc{Negative} & 0.614 \\
carbamazepine & Others on lamatrogine and/or \textbf{carbamazepine}? Hi everyone! & \textsc{Neutral} & 0.990 \\
carbamazepine & This was only the second manic episode I have experienced. My psychiatrist has had a lot of luck with the \textbf{carbamazepine}/Lamotrigine combo. It has done wonders for my depression, and I haven't had a bad manic episode since, maybe periods of mixed (?) episodes. & \textsc{Positive} & 0.998 \\
cariprazine & Anyone have an experience with \textbf{cariprazine} for treatment resistant depression? my doc put my on vraylar after trying several different ssris combined with therapy and having no improvement. & \textsc{Negative} & 0.991 \\
cariprazine & \textbf{cariprazine} dosage for major depression disorder? Hey guys. & \textsc{Neutral} & 0.986 \\
cariprazine & New medicine \textbf{cariprazine}, possible adverse effects so I just recently started on vraylar for treatment resistant depression, and it’s working tremendously I love this medicine. but. & \textsc{Positive} & 0.999 \\
citalopram & Whenever I see the doc he ups the dose to no avail. I have also been on \textbf{citalopram}, escitalopram, fluoxetine and sertraline over the last year, none of which worked out. I am starting to think I could have "treatment resistant depression" because just none of these medications ever make me feel any different. & \textsc{Negative} & 0.996 \\
citalopram & Ssris 1. \textbf{citalopram} 2. lexapro 3. & \textsc{Neutral} & 0.988 \\
citalopram & \textbf{citalopram} I took 10mg for 2 weeks, followed by 20mg for 5 weeks. I noticed a positive in my mood and had anhedonia like feeling eliminated, which was great because I have treatment resistant depression, its been a struggle for years...but my anxiety hasn't been touched. & \textsc{Positive} & 0.998 \\
clomipramine & I went to my partners house and the entire time I felt nothing, nothing good. I felt this way on \textbf{clomipramine} and vilazodone and most other anti depressants. and it’s so stressful im at my wits end with medication. & \textsc{Negative} & 0.992 \\
clomipramine & Have you tried the tcas? specifically, amitriptyline, imipramine and \textbf{clomipramine}? 3. & \textsc{Neutral} & 0.989 \\
clomipramine & I’ve tried like 10 different medications and just last year I was put on a combination of medications that have changed my life. This combination consists of: 300mg of \textbf{clomipramine} and 60mg of Adderall XR These two medications at these dosages have saved my life. I feel so much better now. & \textsc{Positive} & 0.999 \\
clonazepam & However, I have become unmotivated and I find myself sleeping throughout the day. My psychiatrist prescribed me \textbf{clonazepam} and I take 4 mg and sleep for 12+ hours on the weekends. I do not find pleasure in anything. & \textsc{Negative} & 0.994 \\
clonazepam & What should I choose? \&\#x200b; edit: one of my medications is 2mg/d \textbf{clonazepam}, is this a problem for ect? & \textsc{Neutral} & 0.990 \\
clonazepam & *small doses of caffeine have been my friend. and \textbf{clonazepam}/clonazepam, once I got my sleeping under control, helps give me a lovely nap everyday. *I only read about this recently, so it’s a little late for me to try it, but it should be known that patients who are switching from an ssri/snri to parnate, ... & \textsc{Positive} & 0.998 \\
clonidine & From a genetic standpoint, the odds are not in my kids' favor. we tried a low dose of \textbf{clonidine} for anxiety. did not work for the hopelessness and agitation. & \textsc{Negative} & 0.997 \\
clonidine & Or another combination with Wellbutrin? I never tried Intuniv or \textbf{clonidine}... & \textsc{Neutral} & 0.987 \\
clonidine & I now had officially become a drug addict in my opinion. oh, and klonopin (clonazepam) combined with \textbf{clonidine} to be able to sleep at night – no surprise there. I also do not want to leave out that I have had a psychotherapist for 2 years – one of the best things to come out ... & \textsc{Positive} & 0.956 \\
clozapine & Now I have no idea what to do.For context, I have severe, treatment-resistant depression. In the past three years, I’ve tried: • 6 different antidepressants • 5 antipsychotics, including \textbf{clozapine} • Lithium, Vyvanse, tricyclics, and MAOIs • Two full rounds of ECT • Ketamine infusionsNothing worked out. Nothing.The only thing that has made life even slightly ... & \textsc{Negative} & 0.923 \\
clozapine & \textbf{clozapine} makes you fatter than godzilla kinds of statements. or maybe he'd say it of olanzapine and talk about diabetes risk with clozapine. & \textsc{Neutral} & 0.992 \\
clozapine & \textbf{clozapine} makes you fatter than godzilla kinds of statements. or maybe he'd say it of olanzapine and talk about diabetes risk with clozapine. & \textsc{Neutral} & 0.992 \\
deep transcranial magnetic stimulation & At first my brain fog got worse, but it started to clear up as treatment went by. there was one really bad day where the person who placed my \textbf{deep transcranial magnetic stimulation} put it right over a nerve, and it felt like my eye was going to explode. not fun. & \textsc{Negative} & 0.945 \\
deep transcranial magnetic stimulation & Despite of how it comes across, this post in no way shape or form meant to be a promotion for BrainsWay company. It just in the very middle of the point I am trying to make, and I am not aware of any other \textbf{deep transcranial magnetic stimulation} machines which provide stimulation of the entire brain (which ... & \textsc{Neutral} & 0.985 \\
deep transcranial magnetic stimulation & \textbf{deep transcranial magnetic stimulation} But the main difference is that with melancholic depression, where ahedonia is a defining feature, there are no emotions or affect (including no upset) being experienced. The “upset” is only about the fact of your having gotten this condition and it is purely theoretical/analytical, not experienced emotionally but only intellectually (if cognitive abilities are not ... & \textsc{Positive} & 0.795 \\
desipramine & Was very bad for me actually. At one point \textbf{desipramine} worked out for me a little but I didn’t give it long enough, so trying it again. Ritalin is not daily but for mental health emergencies for severe treatment resistant depression, which it has helped in the past. & \textsc{Negative} & 0.478 \\
desipramine & Who knows - not me. Medications I have tried: cipralex, wellbutrin, mylan-mirtazpine, lorazepam, pristiq, sertraline, venlafaxine, cymbalta, prozac, abilify, gabapentin, trintellix, seroquel, trazadone, zopiclone, effexor, nortriptyline, imipramine, \textbf{desipramine}, ketamine infusions (x12), ondansetron, anafranil clomipramine, phenelzine, lithium, zolpidem, citalopram, parnate, viibryd, moclobemide, modafinil, vyvanze, minocycline, cytomel, lithium, amitriptyline, rexulti, concerta, lyrica (pregabalin), adderall (amphetamine XR), luvox (Fluvoxamine), ... & \textsc{Neutral} & 0.972 \\
desipramine & Who knows - not me. Medications I have tried: cipralex, wellbutrin, mylan-mirtazpine, lorazepam, pristiq, sertraline, venlafaxine, cymbalta, prozac, abilify, gabapentin, trintellix, seroquel, trazadone, zopiclone, effexor, nortriptyline, imipramine, \textbf{desipramine}, ketamine infusions (x12), ondansetron, anafranil clomipramine, phenelzine, lithium, zolpidem, citalopram, parnate, viibryd, moclobemide, modafinil, vyvanze, minocycline, cytomel, lithium, amitriptyline, rexulti, concerta, lyrica (pregabalin), adderall (amphetamine XR), luvox (Fluvoxamine), ... & \textsc{Neutral} & 0.972 \\
desvenlafaxine & This is unfortunate because it was working for me otherwise. I switched to \textbf{desvenlafaxine} but it is pretty much causing the same issue. does anyone have any success stories with a different snri that did not because horrible sexual adverse effects?! & \textsc{Negative} & 0.995 \\
desvenlafaxine & Anyone try \textbf{desvenlafaxine} or emsam for trd? hey everyone! & \textsc{Neutral} & 0.989 \\
desvenlafaxine & I couldn’t stay awake for more than 20 minutes while on it. - \textbf{desvenlafaxine} 100 mg/day (current medication): Most effective antidepressant thus far. No adverse effects. & \textsc{Positive} & 0.997 \\
dextroamphetamine & The results came out at nearly the top of the scale and I was diagnosed with fairly severe ADHD. I was prescribed \textbf{dextroamphetamine} and after that Ritalin but have found that both medicines do not relieve my symptoms or help me in any way.. I have read of so many who have 'light bulb' moments ... & \textsc{Negative} & 0.997 \\
dextroamphetamine & So, what really is the difference between vyvanse and \textbf{dextroamphetamine} spansules (and other vyvanse questions)? [x-posted from /r/stims] hi, all. & \textsc{Neutral} & 0.990 \\
dextroamphetamine & Nothing makes me feel better. I was recently diagnosed with adhd and taking \textbf{dextroamphetamine} and that has been going well. my psychiatrist wants to try a last-ditch effort medication before trying spravato. & \textsc{Positive} & 0.996 \\
dextromethorphan bupropion & Running out of options I was on \textbf{dextromethorphan bupropion} for a week. I felt like I was underground, walking in darkness and despair. & \textsc{Negative} & 0.999 \\
dextromethorphan bupropion & Cdphp apparently feels differently. they will not approve spravato because they say “it works on the same receptors and in the same way as \textbf{dextromethorphan bupropion},” and that if I’d like to go on spravato, I need to change to a different po ad. they also (so nicely added) that if I failed auvelity, then I ... & \textsc{Neutral} & 0.988 \\
dextromethorphan bupropion & I was prescribed a small amount of lorazepam as part of an at home migraine rescue cocktail. \textbf{dextromethorphan bupropion}, a new depression med I’ve been trying, has been helping a ton, but it has also caused me to feel a lot more “present” and aware. my anxiety, which I didn’t notice before, is very obvious. & \textsc{Positive} & 0.999 \\
diazepam & They have taught me many techniques and strategies to combat depressive thoughts, suicidal ideation, general anxiousness, anxiety attacks, and insomnia; not one has worked out, and holy shit did I want them to. I have been put on citalopram, fluoxetine, lorazepam, zopiclone, mirtazepine, doxepin, venlafaxine, clonazepam, \textbf{diazepam}, melatonin, quetiapine, lamotrigene, and lithium to help. none of ... & \textsc{Negative} & 0.874 \\
diazepam & *drinking alcohol? no problemo!, you can spend as much money as you want on that, as long as big gov is making some money, pick up some smokes while your at it!, taking some \textbf{diazepam}/alprazolam?* where is your prescription, you are a drug addict, why are you self medicating, you need a fine, criminal record ... & \textsc{Neutral} & 0.990 \\
diazepam & 22. **\textbf{diazepam}** • Familiarity: Quintessential benzodiazepine for anxiety/muscle relaxation. • Antidepressant Potential: None directly; eases anxiety. & \textsc{Positive} & 0.552 \\
doxepin & — either ineffective or made me numb, irritable, more anxious, or even worse physical adverse effects. Antipsychotics: Abilify, Seroquel, Vraylar (initially good, then turned me into a lifeless husk), and Rexulti (see above) Mood stabilizers: Gabapentin, \textbf{doxepin}, Topiramate, Lamictal— nothing worked out, some caused serious adverse effects. Stimulants: Currently on Adderall XR and IR, helps a ... & \textsc{Negative} & 0.966 \\
doxepin & \textbf{doxepin} also hard to find) * medications I did not try yet: Tranylcypromin (Parnate, irreversible MAOI, affects both enzymes in comp to Moclobemide), Clomipramine (but serotonergic drug and ssris/snris never worked out so far), Lamotrigine (Mood Stabilizer, idk if this would be making sense?), Selegiline (MAO-B-I), Amisulpride (low dose should be effective for social anxiety and drive/energy/anhedonia), Tianeptine ... & \textsc{Neutral} & 0.982 \\
doxepin & And I have been productive as hell. my main issues have been insomnia which is sorted out by \textbf{doxepin} 10mg, and a lack of interest in socializing (despite a growing loneliness I'd still rather be working on projects). my fear with switching to nardil is that I will lose the productivy and the progress ive been making ... & \textsc{Positive} & 0.992 \\
duloxetine & Is there a point in trying \textbf{duloxetine}? I have taken many antidepressants and combinations of them for my treatment-resistant depression with no success at all. & \textsc{Negative} & 0.999 \\
duloxetine & Paxil snris 1. \textbf{duloxetine} 2. effexor er atypical antidepressants (adjunct therapy) 1. & \textsc{Neutral} & 0.990 \\
duloxetine & \textbf{duloxetine} + Alcohol = Elevated Heart Rate + Bad Time Hello everyone, Long term lurker first time poster. Been on cymbalta for 2 years 120mg and it’s been extremely effective for treatment resistant depression and GAD. & \textsc{Positive} & 0.999 \\
electroconvulsive therapy & I have undergone soooooo many "treatments" and nothing works. \textbf{electroconvulsive therapy} did not even help. now, my parents will not help pay for me to return to college because they think I am a lost because. & \textsc{Negative} & 0.999 \\
electroconvulsive therapy & Maybe someone here has had \textbf{electroconvulsive therapy} and can speak to this phenomenon. perhaps someone here has researched depression and ect so they ran into research related to this. & \textsc{Neutral} & 0.990 \\
electroconvulsive therapy & Electroconvulsive therapy I’ve been doing \textbf{electroconvulsive therapy} since july and have found it extremely helpful. memory loss is the worst side effect by far. & \textsc{Positive} & 0.999 \\
escitalopram & I’ve been on so many different medications. \textbf{escitalopram} caused my initial weight gain and since switching to multiple others I still can’t lose any weight. I’m so unhappy. & \textsc{Negative} & 0.999 \\
escitalopram & What do you think? as far as I know \textbf{escitalopram} can be prescribed up to 60mg for ocd according to apa guidelines. & \textsc{Neutral} & 0.991 \\
escitalopram & Vyvanse and \textbf{escitalopram} has saved my life! I have treatment resistant depression and anxiety and I’ve tried so many medications. & \textsc{Positive} & 0.999 \\
esketamine & \textbf{esketamine} Not Happy!! It took Spravato for six months and it did absolutely nothing for me. & \textsc{Negative} & 0.999 \\
esketamine & 4. Ketamine therapy in a clinic using the Spravato \textbf{esketamine} nasal spray. 5. & \textsc{Neutral} & 0.990 \\
esketamine & Depressed again I’ve been receiving \textbf{esketamine} treatments for a while now and had been doing great. a real life changer for my treatment resistant depression. & \textsc{Positive} & 1.000 \\
eszopiclone & However, I am doubtful its increased phenelzine's efficacy, whilst the insomnia seems to be a lot worse. I have tried melatonin, diphenhydramine, and my psychiatrist has also prescribed me \textbf{eszopiclone}. none of them work sufficiently well. & \textsc{Negative} & 0.916 \\
eszopiclone & First night I tried 50mg seroquel, second night I tried mirtanza 30mg. now tonight (currently 12:42am) I have had \textbf{eszopiclone} 7.5mg, 6mg alprazolam, 60mg codeine and 15mg diazepam. ( all prescribed medications I keep in a cupboard so none are fake or illegal ). & \textsc{Neutral} & 0.985 \\
eszopiclone & I’m very depressed and rarely have the motivation to do things like brush my hair or get up to drink water if I’m thirsty. so, back to \textbf{eszopiclone}. I am prescribed this for sleep, and it’s helpful for that. & \textsc{Positive} & 0.998 \\
fluoxetine & \textbf{fluoxetine} making me ANGRY and making my ADHD worse. Putting me in a zombie like state. & \textsc{Negative} & 0.999 \\
fluoxetine & More or less clear by now that I have trd with some mood features. I am on 40mg \textbf{fluoxetine}, 2mg abilify and going on lamotragine soon. thinking about throwing inositol into the mix. & \textsc{Neutral} & 0.990 \\
fluoxetine & I’m happy I did because they seem to help. Along with the other supplements, the \textbf{fluoxetine}, and the Adderall XR, my mental health is much better. & \textsc{Positive} & 0.999 \\
fluvoxamine & I have tried 9 different antidepressants for my depression but nothing has brought back my lost interest in things/activities. The antidepressants that I have tried are - Fluoxetine Paroxetine \textbf{fluvoxamine} Escitalopram Sertraline Vilazodone Desvenlafaxine Mirtazapine Bupropion Nothing seems to work. Do I have treatment resistant depression? & \textsc{Negative} & 0.994 \\
fluvoxamine & I talked to my psychiatrist today regarding options because I am extremely drug resistant to anything/any classes, I have literally tried everything. we had \textbf{fluvoxamine} and clomipramine as options, I took the first. I had tried escitalopram/lexapro back in march starting at 10mg, and was encouraged by the sheer amount of positive and non-negative scores on [drugs.com]( ... & \textsc{Neutral} & 0.972 \\
fluvoxamine & I have no adverse effects at all and these medications/doses are currently keeping me in remission. in the past I’ve tried \textbf{fluvoxamine}, sertraline, clomipramine, buspar, lorazepam and wellbutrin. none of these medications worked out besides clomipramine but clomipramine gave me some bad adverse effects like memory problems. & \textsc{Positive} & 0.742 \\
gabapentin & Take 2 to 3 teaspoons of kratom for severe chronic pain. I have taken hydroxyzine, \textbf{gabapentin}, Valium. And a bunch of others that did not help. & \textsc{Negative} & 0.995 \\
gabapentin & Spravato isnt used for anything off label, as far as I know. doet anyone have any anecdotal evidence that they're willkng to sharw about \textbf{gabapentin} use with spravato? also when I was on spravato beforw for roughly 18 months I was on twice a week 84mg almost the entire time. & \textsc{Neutral} & 0.990 \\
gabapentin & It also enabled me to feel empathy, happiness and confidence. \textbf{gabapentin} helps reduce anger, mood swings, and social anxiety. its somewhat like medicinal marijuana but gentler and smoother. & \textsc{Positive} & 0.999 \\
guanfacine & It’s the same feeling as turning a page in a book, you’re met with nothing. It’s almost like the reward system is just fundamentally broken.I’ve tried a bunch of adhd medication (adderall, vyvanse, ritalin, concerta, metadate, dexedrine, strattera, wellbutrin, auvelity, \textbf{guanfacine}) thinking originally it was an ADHD issue but those didn’t help fix this problem ... & \textsc{Negative} & 0.967 \\
guanfacine & Clonidine 3. \textbf{guanfacine} tcas 1. anafrinil others 1. & \textsc{Neutral} & 0.990 \\
guanfacine & Even at work, I’ve made careless mistakes. The only positives so far in these 4 weeks has been from adding \textbf{guanfacine}, which has helped me sleep. There has literally been no change to the Vyvanse. & \textsc{Positive} & 0.964 \\
hydroxyzine & Take 2 to 3 teaspoons of kratom for severe chronic pain. I have taken \textbf{hydroxyzine}, gabapentin, Valium. And a bunch of others that did not help. & \textsc{Negative} & 0.997 \\
hydroxyzine & Bipolar and Borderline were considered but ruled out. medications are as follows: Lithium, Gabapentin, Nortriptylene, prazosin, \textbf{hydroxyzine} (PRN). Here’s the thing, since I met them I’ve noticed that food has a DRAMATIC effect on their mood. & \textsc{Neutral} & 0.984 \\
hydroxyzine & Since the first day I took it, 3 days ago, I’ve felt amazing—extremely and abnormally energetic, talkative, hyper, sleep shorter and still have energy, etc. The other medications I take are Mirtazapine (15), Risperdal (1), Ritalin LA (20), \textbf{hydroxyzine} (50), and propranolol (20). I have been diagnosed with ADHD, treatment-resistant depression, generalized anxiety, and high-functioning ... & \textsc{Positive} & 0.933 \\
imipramine & Treatments that failed for me, in no particular order: 1. \textbf{imipramine} (tca) 2. imipramine + lithium 3. & \textsc{Negative} & 0.993 \\
imipramine & Have you tried the tcas? specifically, amitriptyline, \textbf{imipramine} and clomipramine? 3. & \textsc{Neutral} & 0.989 \\
imipramine & • Familiarity: 9/10 – A classic and well-understood agent in the treatment of depression. • Cost: Low – Generic \textbf{imipramine} is highly affordable. • Effect Size: Powerful – Offers robust antidepressant efficacy, though its adverse effects (sedation, anticholinergic burden, cardiotoxicity) can limit its use. & \textsc{Positive} & 0.921 \\
intermittent theta burst stimulation & Hit rock bottom recently and decided to give neurostimulation a go. first tried 36 sessions of \textbf{intermittent theta burst stimulation}. conplete dud. & \textsc{Negative} & 0.997 \\
intermittent theta burst stimulation & When medications alone aren’t enough, neurostimulation may help some patients. Repetitive transcranial magnetic stimulation and \textbf{intermittent theta burst stimulation} (iTBS) are FDA-approved for individuals with TRD. Electroconvulsive therapy (ECT) is the oldest neurostimulation treatment and is regarded as an effective acute and maintenance intervention in TRD, with preliminary evidence suggesting [noninferiority to acute intravenous ketamine]( & \textsc{Neutral} & 0.715 \\
intermittent theta burst stimulation & Oddly enough, literally the day after I had this plan in place, I spontaneously started spotting — possibly the start of a period — before even starting the medication. Here’s where it gets interesting: I also recently finished a course of \textbf{intermittent theta burst stimulation} TMS (transcranial magnetic stimulation) over a course of 10 consecutive days for ... & \textsc{Positive} & 0.884 \\
isocarboxazid & Most importantly they are thought to be highly effective with trd. common names: emsam, parnate, \textbf{isocarboxazid}, nardil. has anyone had experience with these drugs? & \textsc{Neutral} & 0.988 \\
isocarboxazid & Most importantly they are thought to be highly effective with trd. common names: emsam, parnate, \textbf{isocarboxazid}, nardil. has anyone had experience with these drugs? & \textsc{Neutral} & 0.988 \\
isocarboxazid & I have tried over 35 medications, ECT, ketamine, multiple forms of TMS, ect. The only thing that is helped is \textbf{isocarboxazid}, an MAOI. I have been looking to augment and I am trying to see if lithium is appropriate for my case. & \textsc{Positive} & 0.943 \\
ketamine & Seriously, don’t take Uber X if you can avoid it. My Takeaway \textbf{ketamine} infusions did not work for me. In fact half way through the treatment my depressive symptoms became much much worse. & \textsc{Negative} & 0.999 \\
ketamine & Is that normal? also I saw another thread on here about \textbf{ketamine} addiction? is that possible if we get the infusion through our dr? & \textsc{Neutral} & 0.991 \\
ketamine & These past 3 years have been the best of my life. I knew \textbf{ketamine} was working for me with the first dose (no more ideation at all after the first dose). depression lifted in the first few doses. & \textsc{Positive} & 1.000 \\
lamotrigine & I have been taking \textbf{lamotrigine} for a little over 5 months for treatment resistant depression and anxiety. I have tried 13 other antidepressants and antipsychotics and never had improvements. & \textsc{Negative} & 0.998 \\
lamotrigine & More or less clear by now that I have trd with some mood features. I am on 40mg prozac, 2mg abilify and going on \textbf{lamotrigine} soon. thinking about throwing inositol into the mix. & \textsc{Neutral} & 0.990 \\
lamotrigine & I’ve tried nearly everything while even including multiple different MAOI’S. \textbf{lamotrigine} was finally my saving grace for a year and a half at 250mg!! My entire life, career and relationship changed for the better. & \textsc{Positive} & 1.000 \\
levomilnacipran & \textbf{levomilnacipran} Even other combinations to try that may yield some sort of results. SSRIs: Citalopram – 30 mg once daily Sertraline – 250 mg once daily Paroxetine – 40 mg once daily Escitalopram – 15 mg once daily Fluoxetine – 60 mg once daily SNRIs: Venlafaxine XR – 300 mg once daily Duloxetine – 60 mg ... & \textsc{Neutral} & 0.967 \\
levomilnacipran & \textbf{levomilnacipran} Even other combinations to try that may yield some sort of results. SSRIs: Citalopram – 30 mg once daily Sertraline – 250 mg once daily Paroxetine – 40 mg once daily Escitalopram – 15 mg once daily Fluoxetine – 60 mg once daily SNRIs: Venlafaxine XR – 300 mg once daily Duloxetine – 60 mg ... & \textsc{Neutral} & 0.967 \\
levomilnacipran & \textbf{levomilnacipran} Even other combinations to try that may yield some sort of results. SSRIs: Citalopram – 30 mg once daily Sertraline – 250 mg once daily Paroxetine – 40 mg once daily Escitalopram – 15 mg once daily Fluoxetine – 60 mg once daily SNRIs: Venlafaxine XR – 300 mg once daily Duloxetine – 60 mg ... & \textsc{Neutral} & 0.967 \\
levothyroxine & Is this normal? I have treatment-resistant MDD, and my psychiatrist prescribed me \textbf{levothyroxine} a week ago. I have been feeling even more awful and suicidal since. & \textsc{Negative} & 0.997 \\
levothyroxine & At this medication dosage, I feel like I may even be slightly overmedicated. How did/do these TRD patients take T3 or even \textbf{levothyroxine} without developing a dangerous TSH level? & \textsc{Neutral} & 0.976 \\
levothyroxine & I do not think I am manic now. I recently started \textbf{levothyroxine} we finally got to the bottom of many symptoms linked to hypothyroidism that just did not *quite* have the labs for diagnosis before. I am feeling so much better, not manic just healthy for a first time in a long time. & \textsc{Positive} & 0.997 \\
liothyronine & \textbf{liothyronine} I started taking an ssri in 2016, but while this helped a lot in reducing my anxiety, it made me feel exhausted and completely unmotivated and also made me gain fat really fast. since that time, I have tried countless different medications including two ssris, the newer drugs viibryd and trintillex, a tca, an maoi, ... & \textsc{Negative} & 0.926 \\
liothyronine & Adding \textbf{liothyronine} (liothyronine) as an adjunct to parnate (tranylcypromine). also, should I give more time before I raise the dose? & \textsc{Neutral} & 0.990 \\
liothyronine & The last two months I have been on the **celexa, \textbf{liothyronine}, hydroxizine** cocktail and it is generally working, though I still feel that there could be room for improvement. mainly with the addition of **cytomel** my energy levels are much better and I do not feel the need to sleep all day. & \textsc{Positive} & 0.982 \\
lisdexamfetamine & Since starting at 20mg, I wake up feeling like I got hit by a bus. severe exhaustion, to the point where it feels like my \textbf{lisdexamfetamine} is doing absolutely nothing for me. I feel utterly sedated. & \textsc{Negative} & 0.998 \\
lisdexamfetamine & Wellbutrin and \textbf{lisdexamfetamine} together? I’m not diagnosed with adhd but felt this was where I should post this. & \textsc{Neutral} & 0.989 \\
lisdexamfetamine & Low dose abilify helps a ton with depression and more specifically anhedonia, motivation and apathy. \textbf{lisdexamfetamine} helps a ton with fatigue, motivation and overall mood. this combo has been a life saver. & \textsc{Positive} & 1.000 \\
lithium & My intrusive thoughts are back. I was on \textbf{lithium} for treatment resistant depression . That did not work for me. & \textsc{Negative} & 0.999 \\
lithium & Latuda mood stabilizers 1. \textbf{lithium} ir/er 2. lamictal benzodiazapines 1. & \textsc{Neutral} & 0.989 \\
lithium & I started with moclobemide, which helped a lot, along with pregabalin for the irritability and anxiety. then I added \textbf{lithium} and it helped more. finally I added low dose thyroxine and suddenly I have the energy to feel awake and alive in the morning! & \textsc{Positive} & 0.999 \\
lorazepam & Out of all these medications, only the \textbf{lorazepam} hydroxyzine had any sort of positive effect for me, and they only worked out on the anxiety. Outside of the Lorazepam and hydroxyzine, literally none of these medications helped me in the slightest. & \textsc{Negative} & 0.966 \\
lorazepam & Medical Marijuana 2. \textbf{lorazepam} 3. Hydroxyzine 4. & \textsc{Neutral} & 0.983 \\
lorazepam & After being on this combo for about 8 weeks, I noticed significant benefits. in the meantime I was taking \textbf{lorazepam} up until the medications kicked in. the lexapro helps a ton with anxiety and ocd. & \textsc{Positive} & 0.995 \\
lurasidone & I was having bad anxiety attacks on a daily basis. I never attributed this to the \textbf{lurasidone}, but clearly it wasn't helping. I added effexor 75mg to help combat the anxiety. & \textsc{Negative} & 0.994 \\
lurasidone & Seroqual 5. \textbf{lurasidone} mood stabilizers 1. lithium ir/er 2. & \textsc{Neutral} & 0.989 \\
lurasidone & I was depressed, but I was at least high functioning. I started the \textbf{lurasidone} at the end of september for "treatment resistant depression" ( I have already tried prozac, trintellix, and wellbutrin alone). the first few weeks were awesome. & \textsc{Positive} & 0.995 \\
methylphenidate & I am prescribed \textbf{methylphenidate} for treatment resistant mdd/dysthymia. 15+ standard medications have not worked out and ect left me with horrible memory loss. & \textsc{Negative} & 0.998 \\
methylphenidate & Is anyone in the us, particularly in the northeast, able to get \textbf{methylphenidate} ero or know the current status of the drug? I have trd caused by ptsd. & \textsc{Neutral} & 0.990 \\
methylphenidate & \textbf{methylphenidate} is amazing as someone with asperger's and treatment resistant depression ritalin has been the only drug that has had a positive effect on me. it makes me extroverted and focused at task. & \textsc{Positive} & 1.000 \\
milnacipran & 30. **\textbf{milnacipran} (Savella)** • Familiarity: Approved for fibromyalgia; recognized in some psychiatric circles. • Antidepressant Potential: SNRI activity offers modest mood benefits. & \textsc{Neutral} & 0.954 \\
milnacipran & 30. **\textbf{milnacipran} (Savella)** • Familiarity: Approved for fibromyalgia; recognized in some psychiatric circles. • Antidepressant Potential: SNRI activity offers modest mood benefits. & \textsc{Neutral} & 0.954 \\
milnacipran & 14. **\textbf{milnacipran} (Savella)** • Mechanism: An SNRI that inhibits the reuptake of both serotonin and norepinephrine, thereby modulating pain and mood, particularly in fibromyalgia. • Familiarity: 6/10 – Recognized in fibromyalgia and some psychiatric circles, though not first-line for depression. & \textsc{Positive} & 0.610 \\
mirtazapine & Drastically worsened my anxiety. - \textbf{mirtazapine} 15 mg/day: Severe adverse effects. I couldn’t stay awake for more than 20 minutes while on it. & \textsc{Negative} & 0.999 \\
mirtazapine & Tcas\textbackslash{}* (other than imipramine and clomipramine, due to sri activity) 15. \textbf{mirtazapine} (remeron) 16. trazodone 17. & \textsc{Neutral} & 0.990 \\
mirtazapine & I have been on fluoxetine, which did nothing, sertraline, which sent me batshit insane and landed me in an asylum, amitriptyline, which made me sleep 20 hours a day, and finally \textbf{mirtazapine}. I love mirtazapine. & \textsc{Positive} & 0.999 \\
modafinil & I nap whenever I can. I was prescribed \textbf{modafinil} but it makes me incredibly unhappy and tense for some reason. anyone have luck with anything else? & \textsc{Negative} & 0.999 \\
modafinil & Any other ideas? should I ask about \textbf{modafinil} and, if so, taper the wellbutrin xl qam? for anxiety, perhaps ask for klonopin 2mg over the course of a day along with the gabapentin rx? & \textsc{Neutral} & 0.990 \\
modafinil & Worse than usual for me. yesterday I started the \textbf{modafinil}. I am awake and alert now, I was able to focus and get a ton of work done on a paper that has been stumping me, and most importantly I feel *good.* this is the first med I have taken that I see as actively helping. & \textsc{Positive} & 0.999 \\
nefazodone & I also mentioned maois which she also balked at. has anyone tried \textbf{nefazodone}? comments? & \textsc{Neutral} & 0.987 \\
nefazodone & I also mentioned maois which she also balked at. has anyone tried \textbf{nefazodone}? comments? & \textsc{Neutral} & 0.987 \\
nefazodone & Seldom prescribed antidepressant \textbf{nefazodone} is extremely effective for trd and anxiety. nefazodone is currently the only sndri medication on the market. & \textsc{Positive} & 0.996 \\
nortriptyline & Pdoc told me today I have treatment-resistant depression; suggested ECT. FML Over the past year, I have tried Celexa, Cymbalta, Effexor, Abilify, Ritalin, Latuda, and \textbf{nortriptyline} in various combination. Nothing has really helped. & \textsc{Negative} & 0.992 \\
nortriptyline & Nevertheless, I read in the "the prescriber’s guide to classic mao inhibitors for trd" the following sentence : **in case of insufficient improvement, consider swapping (after washout of 14 days) to phenelzine.** what is the reality ? and if washout is necessary, is there any other class of ad than \textbf{nortriptyline} because I think it is ... & \textsc{Neutral} & 0.988 \\
nortriptyline & Lithium for depression I have treatment-resistant depression. I finally found a med that works and I have been on \textbf{nortriptyline} for 4 years now. it had been helping up until a few months ago. & \textsc{Positive} & 0.999 \\
olanzapine & Did \textbf{olanzapine} because your depression? hi all! & \textsc{Negative} & 0.989 \\
olanzapine & Clozapine makes you fatter than godzilla kinds of statements. or maybe he'd say it of \textbf{olanzapine} and talk about diabetes risk with clozapine. who cares. & \textsc{Neutral} & 0.991 \\
olanzapine & Long story short, I have an unusual sleep disorder where I sleep eight full hours soundly yet do not feel rested. Nothing helped until I stumbled upon \textbf{olanzapine} which worked out wonders overnight, and for the first time in years I woke up refreshed. From what I read, it increases slow wave sleep phases, which apparently in ... & \textsc{Positive} & 0.999 \\
omega 3 & I had tried multivitamins in the past but they did not really help. Neither did \textbf{omega 3}. Maybe Magnesium and Vitamin D worked out faster? & \textsc{Negative} & 0.993 \\
omega 3 & At least 1.5 hours a week. take a multivitamin take \textbf{omega 3} oil capsules look out the window stretch muscles use a vibrator head band for depression, or even rtms, ect for treatment resistant depression smile, because just smiling can apparently create happiness cut toxic people out of your life create flow charts line up ... & \textsc{Neutral} & 0.985 \\
omega 3 & - [ashwagandha]( (life extension, sensoril), has been absolutely crucial for treating my gad/anxiety. - \textbf{omega 3} (important to choose a high quality supplement with low contaminants/oxidization) - *[turmeric/curcimin]( important to research which supplements has high bioavailability if taking as supplement, I use the fresh root simmered with coconut oil/milk and taken with pirpirine) - *tmg (for ... & \textsc{Positive} & 0.760 \\
oxcarbazepine & Lamictal helped me but I am allergic and had to stop. I was then put on \textbf{oxcarbazepine} but that did not help. even off label use of something. & \textsc{Negative} & 0.977 \\
oxcarbazepine & Lamotrigine to \textbf{oxcarbazepine} questions re: depression and hair loss not remotely manic. 54 f, history of major depressive disorder and 2 types of anxiety since adolescence; recently dxed on the autism spectrum (pdd nos) and also complex ptsd. & \textsc{Neutral} & 0.991 \\
oxcarbazepine & I had to completely stop lamictal, I almost died, it was a very scary time for me. I had a pretty incompetent psychiatrist (lost her license and job LMAO) who put me on \textbf{oxcarbazepine} even though she knew my history. and it’s great! & \textsc{Positive} & 0.968 \\
paliperidone & Recovering from Antipsychotics - Seeking regimens to to restore drive and joy I’m about 6 months off Invega (\textbf{paliperidone}), which I was given for a period that now feels like a long, numbed-out survival stretch. During that time, I experienced strong apathy, lack of self-care, no motivation, and near-total anhedonia — especially from things I ... & \textsc{Negative} & 0.915 \\
paliperidone & I can’t take SSRIs. Instead I take Pristiq which is an SNRI and \textbf{paliperidone} for mood stabilization and agitation. I also take concerts for my ADHD. & \textsc{Neutral} & 0.916 \\
paliperidone & I can’t take SSRIs. Instead I take Pristiq which is an SNRI and \textbf{paliperidone} for mood stabilization and agitation. I also take concerts for my ADHD. & \textsc{Neutral} & 0.916 \\
paroxetine & Crying does not release emotions for me, it builds up more. So please do not suggest "more therapy" or "find a different therapist."I have been on Prozac, Lexapro, Zoloft, Effexor, Risperidone, \textbf{paroxetine}, Seroquil, Wellbutrin, and Lithium, and none have worked out. Just... & \textsc{Negative} & 0.997 \\
paroxetine & Luvox 7. \textbf{paroxetine} snris 1. cymbalta 2. & \textsc{Neutral} & 0.988 \\
paroxetine & I of course still have obsessions and compulsions daily, but not nearly as bad as it ever was(I know I will live with this all my life). I have tried many of the antidepressants on the market in the past and have settled with \textbf{paroxetine} for a long time now. it used to really help ... & \textsc{Positive} & 0.998 \\
phenelzine & The worst episode started in Sept 2022. The adverse effects from \textbf{phenelzine} are intolerable (especially insomnia, brain fog) so I intend to stop it after 5 weeks trial if the adverse effects do not improve and a therapeutic benefit is still not apparent. Then I intend to try to start tapering off all the medication. & \textsc{Negative} & 0.996 \\
phenelzine & Are maois the same? my pdoc is looking at \textbf{phenelzine} specifically, but I figure we can generalize, if only to get more responses. & \textsc{Neutral} & 0.990 \\
phenelzine & Sexual dysfunction on \textbf{phenelzine} I’ve been on nardil now for 5 months and it has literally been life saving. I was bed-bound with severe treatment resistant depression and nardil worked out. & \textsc{Positive} & 1.000 \\
pregabalin & I have tried 11 medications. I am currently taking lithium, duloxetine, \textbf{pregabalin}, and aripiprazole. none of these have helped. & \textsc{Negative} & 0.991 \\
pregabalin & Any idea? \textbf{pregabalin} is gabapentinoid = calcium channels, a teeny bit of gaba receptors itself too. tianeptine is opioid = opioid receptors. & \textsc{Neutral} & 0.990 \\
pregabalin & Moclobemide + \textbf{pregabalin} + lithium + thyroxine; this mix gave me my life back! I was in deep, deep treatment-resistant depression (trd) for a long time. & \textsc{Positive} & 0.998 \\
protriptyline & Why? is there something or other about \textbf{protriptyline} that may because liver distress? or is it just tricyclics in general that may because such distress? & \textsc{Negative} & 0.603 \\
protriptyline & My shrink at the cleveland clinic isn't averse to this dose, but he made me have a hepatic panel blood test done today before he will prescribe 60mg/day. he also ordered a \textbf{protriptyline} blood plasma level. why? & \textsc{Neutral} & 0.988 \\
protriptyline & Shrink ordered hepatic panel...and more hello friends, my shrink suggested I take \textbf{protriptyline} a/k/a vivactil, an old tricyclic antidepressant, because I suffer from trd with a lot of anergia and weakness. protriptyline is energizing rather than sedating like the other tricyclics. & \textsc{Positive} & 0.662 \\
psilocybin & I had similar experiences trying \textbf{psilocybin} in the past. should I try a larger dose or come to terms with the fact that psilocybin, like antidepressants, just do not work on me? & \textsc{Negative} & 0.997 \\
psilocybin & From \textbf{psilocybin} and lsd, the next step is often mdma. its like a mix of lsd, shrooms, and caffeine bro, you've already tried all those! & \textsc{Neutral} & 0.990 \\
psilocybin & The results were nothing short of miraculous. I haven't been able to feel joy and calm for decades and one small dose of \textbf{psilocybin} made me feel a calmness, sense of well being and happiness I did not think I could feel. I believe the micro dose had a positive impact with the only adverse effects ... & \textsc{Positive} & 0.999 \\
quetiapine & \textbf{quetiapine} helps my depression a little, what else to try? I am on week 3 of 600 mg of Moclobemide a day and it does not seem to be doing much for my treatment resistant depression. & \textsc{Negative} & 0.998 \\
quetiapine & And clo... I do not remember), abilify, valproic acid, lamotragine, mirtazapine, \textbf{quetiapine}, trintilex, lithium etc. mostly in combination. & \textsc{Neutral} & 0.989 \\
quetiapine & I guess I just needed to vent and ask for opinions. I felt really good on \textbf{quetiapine}, and it completely stopped my manic episode with the first dose. my hallucinations were gone and I felt like a new person. & \textsc{Positive} & 0.999 \\
repetitive transcranial magnetic stimulation & Is \textbf{repetitive transcranial magnetic stimulation} worth it? diagnosed with treatment resistant depression, have wasted years of my life at this point. & \textsc{Negative} & 0.999 \\
repetitive transcranial magnetic stimulation & Do I try to push for an official testing/screening? has anyone ever heard of or tried \textbf{repetitive transcranial magnetic stimulation}? how were you diagnosed with bipolar? & \textsc{Neutral} & 0.990 \\
repetitive transcranial magnetic stimulation & I really think the Deplin is what changed everything for me, but the \textbf{repetitive transcranial magnetic stimulation} undoubtedly helped too. I recommend anyone with treatment-resistant depression ask their doctor about TMS and genetic testing. & \textsc{Positive} & 0.999 \\
risperidone & Crying does not release emotions for me, it builds up more. So please do not suggest "more therapy" or "find a different therapist."I have been on Prozac, Lexapro, Zoloft, Effexor, \textbf{risperidone}, Paxil, Seroquil, Wellbutrin, and Lithium, and none have worked out. Just... & \textsc{Negative} & 0.997 \\
risperidone & Trying TMS On Luvox 150mg and \textbf{risperidone} 1mg. I don’t know if I have true TRD, I haven’t tried any SNRIs or MAOIs, nor have I done anything like ketamine or psilocybin. & \textsc{Neutral} & 0.988 \\
risperidone & Since the first day I took it, 3 days ago, I’ve felt amazing—extremely and abnormally energetic, talkative, hyper, sleep shorter and still have energy, etc. The other medications I take are Mirtazapine (15), \textbf{risperidone} (1), Ritalin LA (20), Hydroxyzine (50), and propranolol (20). I have been diagnosed with ADHD, treatment-resistant depression, generalized anxiety, and high-functioning ... & \textsc{Positive} & 0.907 \\
sam e & -\&gt; proposed remedy (all steps necessary): a) you must keep a medication diary. b) you must keep your doctor in the loop if you plan on taking anything alongside your prescribed medication, such as \textbf{sam e}. if you're taking an maoi, you must take particular care with what else you ingest. & \textsc{Neutral} & 0.988 \\
sam e & -\&gt; proposed remedy (all steps necessary): a) you must keep a medication diary. b) you must keep your doctor in the loop if you plan on taking anything alongside your prescribed medication, such as \textbf{sam e}. if you're taking an maoi, you must take particular care with what else you ingest. & \textsc{Neutral} & 0.988 \\
sam e & -\&gt; proposed remedy (all steps necessary): a) you must keep a medication diary. b) you must keep your doctor in the loop if you plan on taking anything alongside your prescribed medication, such as \textbf{sam e}. if you're taking an maoi, you must take particular care with what else you ingest. & \textsc{Neutral} & 0.988 \\
selegiline & Quit \textbf{selegiline} 6mg after experiencing severe chest pain - what now? the pain was getting progressively worse and made it difficult to breathe, so I did not want to risk continuing the medication. & \textsc{Negative} & 0.995 \\
selegiline & I have read two. Anyone like \textbf{selegiline}, what do you do for ocd? Thank you? & \textsc{Neutral} & 0.990 \\
selegiline & I am diagnosed with Treatment-Resistant Depression, Obsessive-Compulsive Disorder and Generalized Anxiety Disorder. \textbf{selegiline} has been nothing short of amazing for my depression and intrusive thoughts. My only real low cost options seem to be Tranylcypromine and Phenelzine. & \textsc{Positive} & 0.999 \\
sertraline & I cannot cope anymore. I take 100mg \textbf{sertraline} since 2017 and tried multiple therapies but nothing helped. I cannot take another of these heavy episodes. & \textsc{Negative} & 0.998 \\
sertraline & \textbf{sertraline} (200mg approx?) actually does. So even tho its dangerous doc thinks of combining effexor with zoloft. & \textsc{Neutral} & 0.990 \\
sertraline & I took \textbf{sertraline} in my early 20s which really helped my depression and anxiety. I would get intrusive thoughts of harm towards myself or others, I would worry about what if I jumped off a bridge etc zoloft really helped out for a good 8 years. & \textsc{Positive} & 0.999 \\
thyroid & Was prescribed \textbf{thyroid} hormones and I feel a lot worse. Is this normal? & \textsc{Negative} & 0.999 \\
thyroid & 11. have you tried thyroxine (synthetic \textbf{thyroid} hormone) and minocycline (antibiotic) as adjunctive agents in combination with your antidepressant regime? no one is allowed to give up unless you have tried lithium and one to eleven above. & \textsc{Neutral} & 0.989 \\
thyroid & Moclobemide + pregabalin + lithium + \textbf{thyroid}; this mix gave me my life back! I was in deep, deep treatment-resistant depression (trd) for a long time. & \textsc{Positive} & 0.998 \\
topiramate & — either ineffective or made me numb, irritable, more anxious, or even worse physical adverse effects. Antipsychotics: Abilify, Seroquel, Vraylar (initially good, then turned me into a lifeless husk), and Rexulti (see above) Mood stabilizers: Gabapentin, Doxepin, \textbf{topiramate}, Lamictal— nothing worked out, some caused serious adverse effects. Stimulants: Currently on Adderall XR and IR, helps a ... & \textsc{Negative} & 0.964 \\
topiramate & I also read once that if severe agitation is present with depression, it is assumed psychosis, but I dont know why or if that applies in my case. someone online once mentioned to me that rage sits in your temporal lobes, you need \textbf{topiramate}. I never got a chance to have a clarification what that ... & \textsc{Neutral} & 0.989 \\
topiramate & Been on a couple snris and no real improvement. the most effective regimen I was on was \textbf{topiramate} + klonopin, topamax off label for mood stabilization. it worked out great for about 3 years, and then it made things way worse. & \textsc{Positive} & 0.915 \\
tranylcypromine & \textbf{tranylcypromine} is worsening my symptoms. Could you please give me some reassurance? & \textsc{Negative} & 0.998 \\
tranylcypromine & Adding t3 (liothyronine) as an adjunct to \textbf{tranylcypromine} (tranylcypromine). also, should I give more time before I raise the dose? & \textsc{Neutral} & 0.990 \\
tranylcypromine & But I want to share my experience, maybe it is worth a try for people who tried already everything else without response. I have been taking \textbf{tranylcypromine} now for 8 years in various dosages (up to 80 mg). it is a reliable chemical tool, that gave me back control over my life. & \textsc{Positive} & 0.999 \\
trazodone & She had recently added \textbf{trazodone} to help with my anxiety and insomnia but the depression got worse. I’m on day six of tapering off the trazodone and it is hell. & \textsc{Negative} & 0.996 \\
trazodone & And also told him I don’t want to get this because I’m trying to self medicate and abuse it. and let him know that I have copious amounts of klonopin, \textbf{trazodone}, vyvanse in the house if I wanted to abuse medication and self medicate but I don’t touch anything. it really pisses me off. & \textsc{Neutral} & 0.988 \\
trazodone & 3 doses taken. \textbf{trazodone} - Successful remission in most anxiety symptoms. Taking since 2019. & \textsc{Positive} & 0.999 \\
trimipramine & \textbf{trimipramine} also hard to find) * medications I did not try yet: Tranylcypromin (Parnate, irreversible MAOI, affects both enzymes in comp to Moclobemide), Clomipramine (but serotonergic drug and ssris/snris never worked out so far), Lamotrigine (Mood Stabilizer, idk if this would be making sense?), Selegiline (MAO-B-I), Amisulpride (low dose should be effective for social anxiety and drive/energy/anhedonia), Tianeptine ... & \textsc{Neutral} & 0.984 \\
trimipramine & \textbf{trimipramine} also hard to find) * medications I did not try yet: Tranylcypromin (Parnate, irreversible MAOI, affects both enzymes in comp to Moclobemide), Clomipramine (but serotonergic drug and ssris/snris never worked out so far), Lamotrigine (Mood Stabilizer, idk if this would be making sense?), Selegiline (MAO-B-I), Amisulpride (low dose should be effective for social anxiety and drive/energy/anhedonia), Tianeptine ... & \textsc{Neutral} & 0.984 \\
trimipramine & \textbf{trimipramine} also hard to find) * medications I did not try yet: Tranylcypromin (Parnate, irreversible MAOI, affects both enzymes in comp to Moclobemide), Clomipramine (but serotonergic drug and ssris/snris never worked out so far), Lamotrigine (Mood Stabilizer, idk if this would be making sense?), Selegiline (MAO-B-I), Amisulpride (low dose should be effective for social anxiety and drive/energy/anhedonia), Tianeptine ... & \textsc{Neutral} & 0.984 \\
valproate & About two weeks ago my doctor prescribed me 500mg of sodium \textbf{valproate} alongside my regular dose of 40mg of fluoxetine (prozac). I do not have bipolar as far as I know so I am not really the typical patient that gets prescribed sodium valproate. & \textsc{Negative} & 0.987 \\
valproate & He basically gave me the go-ahead to try it because he said we’ve done so many other medications, lots of antidepressants. I’ve tried \textbf{valproate} I’ve tried Lamictal. If you name the med I’ve had a go barring the trycyclics. & \textsc{Neutral} & 0.987 \\
valproate & So thankful I wasn't alone at home.a few days later, I found that \textbf{valproate} (divalproex, 500mg) gave a better effect than vigabatrin. the combination (tcp, olanzapine, valproate, quetiapine, methylphenidate) allowed me to survive until I got a couple more bottles of greenstone nardil sent me by another colleague. & \textsc{Positive} & 0.936 \\
venlafaxine & It was the worst experience I have ever had with medications. I fucking hate \textbf{venlafaxine}. my depression relapsed to the worst it is ever been, presumably because of it. & \textsc{Negative} & 0.999 \\
venlafaxine & Cymbalta 2. \textbf{venlafaxine} er atypical antidepressants (adjunct therapy) 1. wellbutrin 2. & \textsc{Neutral} & 0.990 \\
venlafaxine & I previously tried Paxil and Zoloft. \textbf{venlafaxine} worked out very well and had minimal adverse effects. I had 36mg of Concerta added Sept 2015 for sleepiness and trouble concentrating. & \textsc{Positive} & 0.999 \\
vilazodone & Nope.. tried \textbf{vilazodone} at a very low dose, half the minimum therapeutic dose, in fact... ended up suicidal and the depression went off the charts.. & \textsc{Negative} & 0.995 \\
vilazodone & Current medications: desvenlafaxine er 100mg (generic pristiq), generic adderall xr (20mg), and generic adderall ir (10 mg twice per day). past medications: -antidepressants: venlafaxine, \textbf{vilazodone}, bupropion, prozac, zoloft, lexapro, nortripytilene, and emsam patch. -also tried both tms and ect. & \textsc{Neutral} & 0.984 \\
vilazodone & I have treatment resistant depression and have been on several SSRIs and SNRIs. My most recent combo is 40mg \textbf{vilazodone} and 1.5mg Vraylar which I have been taking for around six months now. Almost immediately recognized a difference which is unheard of for me-- no cyclical negative thoughts, no panic attacks or random crying outbursts. & \textsc{Positive} & 0.995 \\
vortioxetine & I’ve tried 5 other anti depressants that didn’t work for me. Wondering if folks with treatment resistant depression have seen \textbf{vortioxetine} work? I’ve been on for about 8 weeks now and I’m not seeing anything. & \textsc{Negative} & 0.998 \\
vortioxetine & \textbf{vortioxetine}/Trintellix tollerance? Hi all! & \textsc{Neutral} & 0.988 \\
vortioxetine & My med combo is working great \textbf{vortioxetine} + Bupropion + Vyvanse. !! & \textsc{Positive} & 0.999 \\
zaleplon & Olanzapine ODT 16. \textbf{zaleplon} 17. Clonazepam ODT 18. & \textsc{Neutral} & 0.982 \\
zaleplon & Olanzapine ODT 16. \textbf{zaleplon} 17. Clonazepam ODT 18. & \textsc{Neutral} & 0.982 \\
zaleplon & Olanzapine ODT 16. \textbf{zaleplon} 17. Clonazepam ODT 18. & \textsc{Neutral} & 0.982 \\
ziprasidone & In the past, antidepressants have helped temporarily but they always eventually stop working. I have been on a lot of medications used for bipolar disorder (zyprexa, risperdal, lithium, seroquel, lamictal, abilify, \textbf{ziprasidone}) but because I do not experience mania I have never found them helpful for treating my symptoms, and a lot of them caused intolerable side ... & \textsc{Negative} & 0.968 \\
ziprasidone & Combo medications of an SNRI and an antipsychotic come to mind but I do not know. Maybe duloxetine and \textbf{ziprasidone}? TL;DR: Advice for combo medications on difficult to treat depression. & \textsc{Neutral} & 0.984 \\
ziprasidone & My depression is back and as ugly as ever unless I take my dex now even after upping emsam to 9mg. my (and my psych’s) next thought is either parnate, symbyax, or adding \textbf{ziprasidone} back as atypical antipsychotics have worked out well for my (mostly anergic and anhedonic) depression in the past. & \textsc{Positive} & 0.877 \\
zolpidem & I can’t take insomnia anymore. Why \textbf{zolpidem} doesn’t help me. I’m so terrified nth or any somewhat augmenting Nardil with get a boost and I have two teenagers. & \textsc{Negative} & 0.993 \\
zolpidem & Im not even sure if this is allowed, but I figured wtf I will check to see if you guys know or have some feedback. do you think a lot \textbf{zolpidem} cr 12.5mg mixed with a lot of 100mg wellbutrin will achieve self deletion? & \textsc{Neutral} & 0.988 \\
zolpidem & What is the point of anything, better off dead, whatever. narcolepsy diagnosis came as no surprise, \textbf{zolpidem} was a miracle. I could sleep at night. & \textsc{Positive} & 0.999 \\
zuranolone & Could it be of harm? You bet.” Other investigational treatments for TRD include lithium, thyroid hormone, buspirone, L-methylfolate, S-adenosylmethionine, and various anti-inflammatory agents such as cyclooxygenase-2 inhibitors, minocycline, statins, and tumor necrosis factor–alpha antagonists, as well as \textbf{zuranolone} and the dextromethorphan-bupropion combination.” & \textsc{Neutral} & 0.974 \\
zuranolone & Could it be of harm? You bet.” Other investigational treatments for TRD include lithium, thyroid hormone, buspirone, L-methylfolate, S-adenosylmethionine, and various anti-inflammatory agents such as cyclooxygenase-2 inhibitors, minocycline, statins, and tumor necrosis factor–alpha antagonists, as well as \textbf{zuranolone} and the dextromethorphan-bupropion combination.” & \textsc{Neutral} & 0.974 \\
zuranolone & Could it be of harm? You bet.” Other investigational treatments for TRD include lithium, thyroid hormone, buspirone, L-methylfolate, S-adenosylmethionine, and various anti-inflammatory agents such as cyclooxygenase-2 inhibitors, minocycline, statins, and tumor necrosis factor–alpha antagonists, as well as \textbf{zuranolone} and the dextromethorphan-bupropion combination.” & \textsc{Neutral} & 0.974 \\
\end{longtable}
\endgroup

\end{document}